\documentclass[sigconf]{acmart}
\AtBeginDocument{%
  }

\acmConference[]{}{}{}
\usepackage{multirow}
\usepackage{balance}
\usepackage{makecell}
\usepackage{subcaption}

\setcopyright{none}
\makeatletter
\@ACM@noshowfoottrue
\makeatother
\begin{document}

%%
%% The "title" command has an optional parameter,
%% allowing the author to define a "short title" to be used in page headers.
\title{Differential Contrastive Training for Gaze Estimation}

%%
%% The "author" command and its associated commands are used to define
%% the authors and their affiliations.
%% Of note is the shared affiliation of the first two authors, and the
%% "authornote" and "authornotemark" commands
%% used to denote shared contribution to the research.
\author{Lin Zhang}
\orcid{0009-0002-1824-2270}
\affiliation{
  \institution{Beijing Key Laboratory of Traffic Data Mining and Embodied Intelligence}
  \department{Beijing Jiaotong University}
  \city{Beijing}
  \country{China}
}
\email{23125286@bjtu.edu.cn}

\author{Yi Tian}
\authornote{Corresponding Author.}
\orcid{0000-0001-6054-7970}
\affiliation{
  \institution{Beijing Key Laboratory of Traffic Data Mining and Embodied Intelligence}
  \department{Beijing Jiaotong University}
  \city{Beijing}
  \country{China}
}
\email{tianyi@bjtu.edu.cn}

\author{Xiyun Wang}
\orcid{0009-0001-2660-8749}
\affiliation{
  \institution{Beijing Key Laboratory of Traffic Data Mining and Embodied Intelligence}
  \department{Beijing Jiaotong University}
  \city{Beijing}
  \country{China}
}
\email{24120389@bjtu.edu.cn}

\author{Wanru Xu}
\orcid{0000-0003-2206-5051}
\affiliation{%
  \institution{School of Computer Science and Technology}
  \institution{Beijing Jiaotong University}
  \city{Beijing}
  \country{China}
}
\email{xwanru@bjtu.edu.cn}

\author{Yi Jin}
\orcid{0000-0001-8408-3816}
\affiliation{%
  \institution{School of Computer Science and Technology}
  \institution{Beijing Jiaotong University}
  \city{Beijing}
  \country{China}
}
\email{yjin@bjtu.edu.cn}

\author{Yaping Huang}
\orcid{0000-0002-4465-372X}
\affiliation{%
  \institution{Beijing Key Laboratory of Traffic Data Mining and Embodied Intelligence}
  \department{Beijing Jiaotong University}
  \city{Beijing}
  \country{China}
}
\email{yphuang@bjtu.edu.cn}

%%
%% By default, the full list of authors will be used in the page
%% headers. Often, this list is too long, and will overlap
%% other information printed in the page headers. This command allows
%% the author to define a more concise list
%% of authors' names for this purpose.
\renewcommand{\shortauthors}{Lin Zhang et al.}

%%
%% The abstract is a short summary of the work to be presented in the
%% article.
\begin{abstract}
The complex application scenarios have raised critical requirements for precise and generalizable gaze estimation methods. Recently, the pre-trained CLIP has achieved remarkable performance on various vision tasks, but its potentials have not been fully exploited in gaze estimation. 
In this paper, we propose a novel Differential Contrastive Training strategy, which boosts gaze estimation performance with the help of the CLIP. 
Accordingly, a Differential Contrastive
Gaze Estimation network (DCGaze) composed of a Visual Appearance-aware branch and a Semantic Differential-aware branch is introduced. 
The Visual Appearance-aware branch is essentially a primary gaze estimation network and it incorporates an Adaptive Feature-refinement Unit (AFU) and a Double-head Gaze Regressor (DGR), which both help the primary network to extract informative and gaze-related appearance features. 
Moreover, the Semantic Difference-aware branch is designed on the basis of the CLIP's text encoder to reveal the semantic difference of gazes. 
This branch could further empower the Visual Appearance-aware branch with the capability of characterizing the gaze-related semantic information.  
Extensive experimental results on four challenging datasets over within and cross-domain tasks demonstrate the effectiveness of our DCGaze. \textcolor{blue}{The code is available at \textit{\url{https://github.com/LinZhang-bjtu/DCGaze}}}. 
\end{abstract}

\maketitle

\begin{figure}[h]
 \centering
   \includegraphics[width=0.9\linewidth]{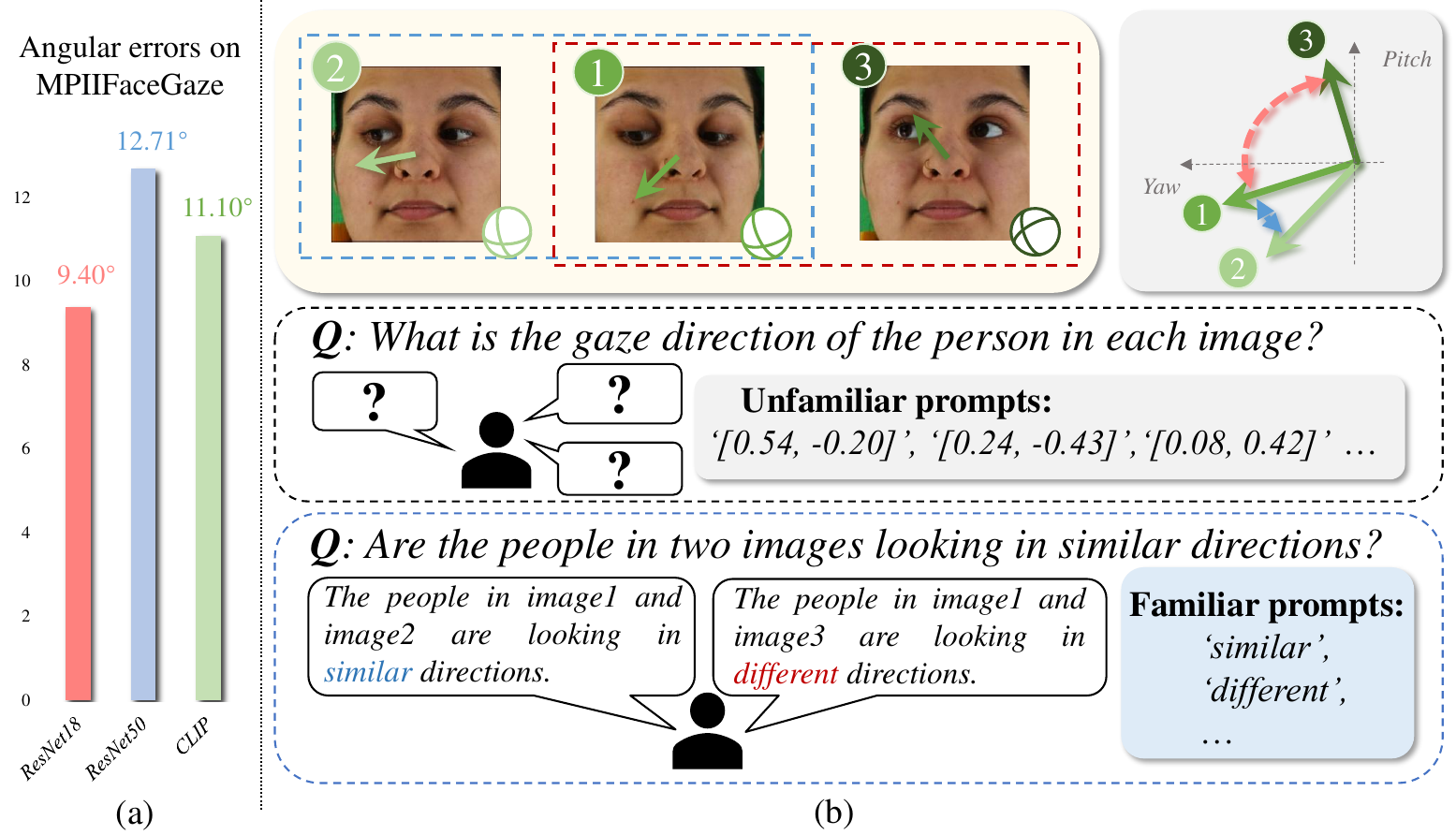}
   \vspace{-10pt}
   \caption{(a) The gaze errors of ResNet18, ResNet50 and CLIP on MPIIFaceGaze dataset. We evaulate ResNet18 and ResNet50 on the cross-domain task $\mathcal{D}_E  \rightarrow \mathcal{D}_M$, which indicates training on ETH-XGaze and testing on MPIIFaceGaze.  
   (b) Comparing with generating a unique sentence for each image, it is more easier to describe the gaze difference of a pair of images.}
   \label{fig:1} 
\end{figure}

\section{Introduction}
\label{sec:intro}
Gaze estimation aims to predict a 2D gaze position or a 3D gaze direction of a specified user given its facial or eye image. As an important task in the field of computer vision, it has been extensively utilized in various people-related researches, such as virtual reality~\cite{Burova2020UtilizingVA,gazeVR}, autonomous driving~\cite{Drivers,chengWhatYouSee2024}, etc. In those real scenarios, the users’ identities and their surrounding environments are changeable and complicated, which puts a heavy responsibility on the accuracy and generalization of gaze estimation models. In other words, the estimation models that are optimized via training users should adapt to disjointed new subjects accurately.

Most recently, the pre-trained Visual-Language models represented by CLIP~\cite{CLIP} have demonstrated impressive performance on various perception tasks, like image classification~\cite{chertiReproducibleScalingLaws2023}, semantic segmentation~\cite{ZegCLIP}, depth estimation~\cite{DepthCLIP,WorDepth} and others~\cite{CLIPHand3D}. CLIP trains an image encoder and a text encoder jointly by a contrastive loss in the embedding space between both modalities, which endows it with powerful representation capabilities by characterizing both visual and semantic information of a target object. Naturally, we are spontaneously curious about the issue: \textit{\textbf{can CLIP understand human gaze? }} 
To explore this concern, we make a simple early attempt. Inspired by~\cite{DepthCLIP}, we define four gaze bins that align with the gaze directions of  $[0,\frac{\pi}{2}]$, $[0,-\frac{\pi}{2}]$, $[\frac{\pi}{2},0]$ and $[-\frac{\pi}{2},0]$, respectively. And their semantic descriptions are designated as a prescribed prompt `A photo of a face gazing \{\textit{gaze direction}\}.’ where gaze direction includes [`up’, `down’, `left’, `right’]. Then the gaze direction of an arbitrary facial image could be obtained via linearly combining the multi-bin gaze values according to the language-image similarities between its visual embedding and semantic embeddings of all the gaze bins (details refer to supplementary materials). Surprisingly, even without fine-tuning on any exclusive gaze estimation datasets, the CLIP achieves promising performance comparable to ResNet18 and ResNet50 that trained with cross-domain samples (Figure~\ref{fig:1}). From the results, we thus draw a rough conclusion that \textit{\textbf{the CLIP could perceive human gazes subtly, but its potentials have not been fully exploited}}.

We theoretically analyze the possible reasons hindering the CLIP from reaching its full potentials in perceiving gaze directions. On the one hand, CLIP has never seen standard `face (image)-gaze (text)’ pairs in its pre-training stage. This makes it difficult for CLIP to understand the meaning of gaze directions,e.g., [0.54 (pitch), -0.20 (yaw)]. On the other hand, unlike the finite and discrete outputs in typical classification and segmentation tasks, gaze labels are distributed in an infinite and continuous space. Thus, it is difficult to design a set of gaze-guided text prompts that can fittingly align with each facial image. In other words, the challenges in connecting text prompts with gaze labels obstruct the extraction of gaze-related semantic information. 

Recently, several studies \cite{CLIPGaze,GazeCLIP} have attempted to take advantage of CLIP to predict gaze directions. The notable CLIP-Gaze \cite{CLIPGaze} aims to purify gaze-relevant features by pushing away the extracted gaze features from gaze-irrelevant semantic embeddings obtained via the text encoder of CLIP.
Essentially, it employs the text encoder to generate gaze distractors rather than describe the gaze itself, which remains unable to solve the above issues. %\looseness = -1

To address the aforementioned challenges, in this paper, we put forward an innovative and straightforward \textbf{Differential Contrastive Training (DCTrain)} strategy to fully evoke CLIP’s potentials in gaze estimation. 
Commonly, although it is impractical to describe the continuous gaze direction of each facial image in language, it is easier to semantically distinguish the gaze difference between a sample pair. For example, as shown in Figure~\ref{fig:1}, there are three facial images with gaze directions of $[0.54, -0.20]$, $[0.24, -0.43]$, $[0.08, 0.42]$, respectively. Obviously, we could not generate a unique sentence for each image to explain their detailed gaze directions. While we could clearly draw the conclusions that `the people in \textit{image1} and \textit{image2} are looking in  \textbf{\textit{similar}} directions' and `the people in \textit{image1} and \textit{image3} are looking in  \textbf{\textit{different}} directions.’ Motivated by the common observations, instead of identifying gaze direction of single facial image, our Differential Contrastive Training strategy is designed to leverage CLIP for unveiling gaze semantic differences of image pairs. 
Innovatively, this strategy could simplify the target texts that the CLIP needs to understand by replacing its unfamiliar phrases, e.g., `$[0.54, -0.20]$'  with familiar ones, e.g., `similar' or `different' (Figure~\ref{fig:1}). Moreover, the continuous gaze labels are transformed into discrete similarity levels of a pair of gazes.
Thus, the above two challenges are expected to be resolved under our Differential Contrastive Training strategy. \looseness = -1

% This cleverly address the challenges that hinder the applications of CLIP in gaze estimation

Accordingly, a \textbf{Differential Contrastive Gaze Estimation network (DCGaze)} is proposed, which consists of a Visual Appearance-aware branch and a Semantic Difference-aware branch. The former is essentially a primary gaze estimation network, which extracts the appearance feature of each facial image and then regresses it into a final gaze direction. We employ the remarkable CNN-Transformer architecture as the basic structure of this branch. To ensure that the extracted appearance features are informative and gaze-related, we propose an Adaptive Feature-refinement Unit (AFU) and a Double-head Gaze Regressor (DGR). 
The former is based on the image encoder of the pre-trained CLIP, which aims to dynamically enhance the gaze-related contents contained in extracted appearance features via thoroughly exploring its relationships with the prior appearance features of CLIP.
The Double-head Gaze Regressor is introduced to mitigate the overfitting issues caused by the conventional MLP-based regressor. It incorporates an extra regressor head with random masking policy to prevent itself from excessively focusing on several feature dimensions. The later branch is built upon the text encoder of CLIP, which could be plug and play to arbitrary gaze estimation models. Firstly, a series of differential gaze prompts are designed, which reveal the relationships of gaze directions between two facial images in language (Table ~\ref{tab:7}). And then, the connections between the semantic embeddings of these prompts and the appearance features of image pairs captured via the Visual Appearance-aware branch could be established to realize an `image-language’ contrastive learning. To be specific, the `image’ refers to the concatenated appearance features of selected image pairs. The `language’ refers to the sentence that describes their similarity in terms of true gaze directions. Therefore, by semantically identifying the gaze difference of sample pairs with the help of the Differential Contrastive Training, we can fully leverage the semantic representation ability of CLIP to facilitate the primary gaze estimation model to characterize semantic gaze-related information.

The main contributions of our paper are as follows:

$\bullet$ We develop the potentials of CLIP in boosting gaze estimation performance with a novel \textbf{Differential Contrastive Training} strategy. Innovatively, we propose to establish the connections between languages and gaze differences of image pairs, which could provide new ideas for analogous quantified/regression tasks. 

$\bullet$ We propose a novel \textbf{Differential Contrastive Gaze Estimation network}, which consists of a Visual Appearance-aware branch and a Semantic Difference-aware branch. The former is designed to extract more robust appearance features and build a generalized regressor. 
The latter improves the ability to characterize the semantic gaze-related information of the primary gaze network by unveiling gaze semantic difference.
% designing multiple text prompts to represent the text differences in image pairs.
    
$\bullet$ Our network achieves competitive performance over both within-domain and cross-domain tasks on MPIIFaceGaze, EyeDiap, Gaze360 and ETH-XGaze datasets. 
    
    % Besides, serving as a plug-and-play plugin, the XX branch along with our \textbf{Differential Contrastive Training} strategy brings performance improvements to various basic gaze estimation models.

\begin{figure*}[t]
  \centering
  %\fbox{\rule{0pt}{2in} \rule{0.9\linewidth}{0pt}}
   \includegraphics[width=0.9\linewidth]{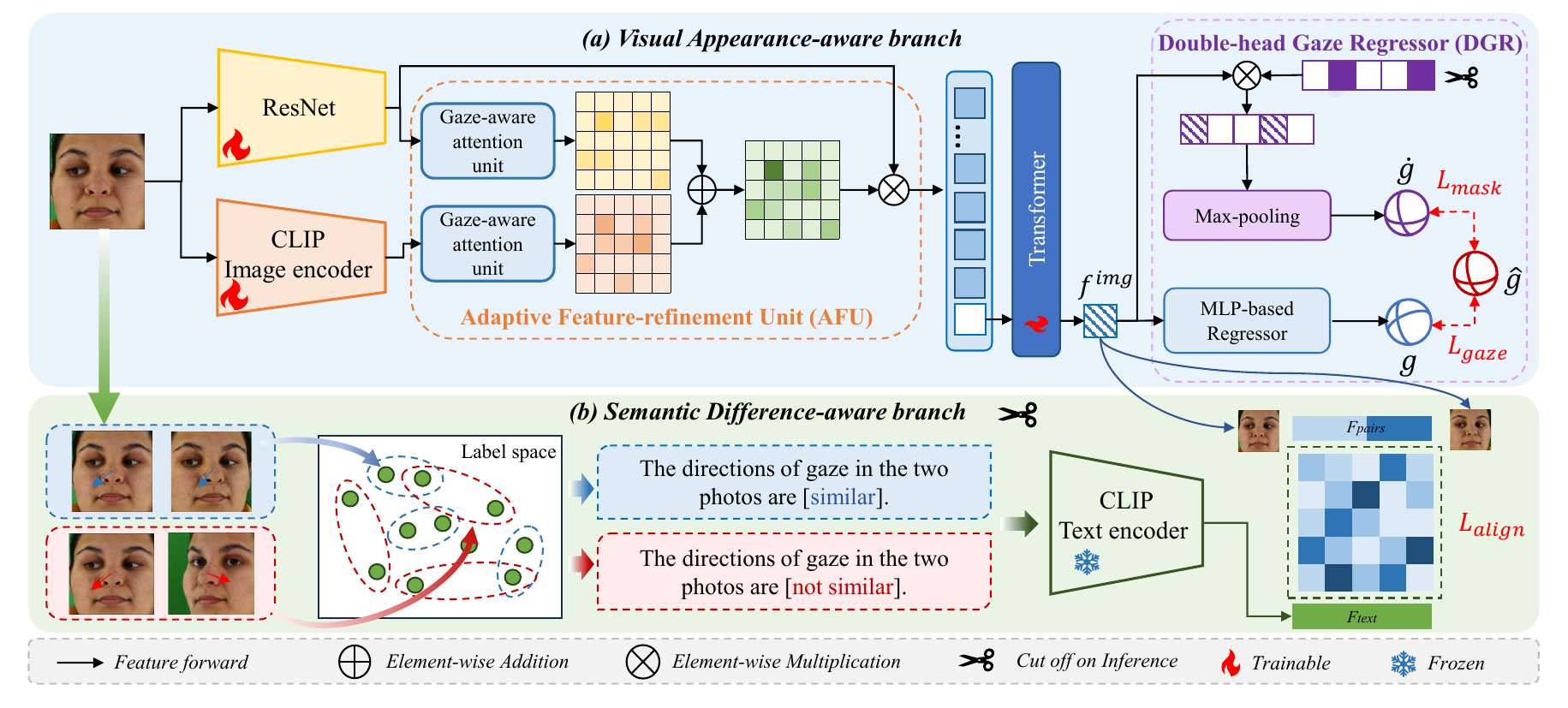}
   \vspace{-10pt}
   \caption{Framework of our proposed DCGaze. It consists of two branch: Visual Appearance-aware branch and Semantic Difference-aware branch. In  Visual Appearance-aware branch, the AFU is designed to adaptively enhance the gaze-related contents of primary appearance features with the guidance of the prior appearance features of CLIP's image encoder. The DGR maps those enhanced features to final gazes via two regression heads. The Semantic Difference-aware branch selects several image pairs from sample batch and gives each of them a textual prompt that describes their gaze differences. Then, the prompts are fed into CLIP's text encoder to capture text embeddings, which are then aligned with enhanced appearance features obtained by our Visual Appearance-aware branch.}
   \vspace{-10pt}
   \label{fig:2}
\end{figure*}

\section{Related Work}
\textbf{Gaze estimation.}
With the recent developments in deep learning, appearance-based gaze estimation methods have become the mainstream. Researchers explored various CNN-based architectures~\cite{FullFace,CA-Net,Dilated-Net} to build the mapping between facial images and gaze directions. Zhang et al.~\cite{AppearancebasedGazeEstimation} firstly proposed a CNN-based gaze estimation network and the well-known MPIIFaceGaze dataset. With the introduction of Transformer~\cite{Transformer}, the Vision Transformer-based structures have started to be applied to gaze estimation and achieved promising performance (e.g., GazeTR~\cite{GazeTR}, oh et al. ~\cite{oh}, SUGE~\cite{SUGE}).
In the recent years, researchers dedicated themselves to inventing generalized gaze estimation methods, which would show robust performance on unseen users. Several methods leveraged a gaze redirection strategy to extend the datasets for generalized gaze estimation~\cite{DeepWarp,GFAL}. Cheng et al.~\cite{PureGaze} introduced a method of purifying gaze features to improve the network’s generalization. Besides, some methods based on contrastive learning~\cite{ContrastiveRegressionDomain}and uncertainty learning~\cite{UncertaintyModelingGaze,SUGE} also demonstrated remarkable generalizability. In this paper, we leverage CLIP to improve discriminability and generalization of the gaze estimation network. 

\textbf{Pre-trained Vision-language Models.}
Recently, the CLIP~\cite{CLIP} trained on large-scale image-text pairs have attracted increasing attentions. Because of its powerful visual and semantic representation capabilities, CLIP has been transferred to various vision tasks, such as objection detection~\cite{CLIPGap}, semantic segmentation~\cite{ZegCLIP}, and others~\cite{Grounded,CLIPHOI,CLIPSS}. Especially, CLIP also shows surprising capacities on quantified vision tasks, e.g., depth estimation~\cite{DepthCLIP,WorDepth}, 3D hand poses estimation~\cite{CLIPHand3D}, etc. Moreover, researchers were attempting to take advantages of the pre-trained CLIP to predict gaze directions. Wang et al.~\cite{GazeCLIP} designed a linguistic description generator to produce text signals with coarse gaze directional cues. And then a cross-attention condenser was designed to finely recalibrate the visual and text representations, enhancing the learning quality of gaze features. Yin et al.~\cite{CLIPGaze} designed a feature separation loss by employing CLIP text encoder to generate gaze distractors from diverse language descriptions, which aims at purifying the gaze-relevant feature via pushing away it from gaze-irrelevant features. In this paper, we adapt CLIP to gaze estimation with an innovative collaborative enhancing strategy, in which the CLIP is regarded as an assistance to enhance the obtained gaze features.
\vspace{-5pt}

\section{Methodology}
\label{sec:method}
To fully activate the potentials of CLIP to perceive gaze directions, we propose a novel Differential Contrastive Training (DCTrain) strategy, in which the CLIP is leveraged for unveiling gaze semantic differences of image pairs. Accordingly, we propose a \textbf{Differential Contrastive Gaze Estimation network (DCGaze)} as shown in Figure~\ref{fig:2} which consists of a Visual Appearance-aware branch and a Semantic Difference-aware branch. The former is essentially a primary gaze estimation network, while the latter is to encourage the Visual Appearance-aware branch to extract gaze-related features with the help of image-language contrastive learning.
\vspace{-5pt}
\subsection{Visual Appearance-aware Branch}
The Visual Appearance-aware branch is composed of a Basic Appearance Feature Extractor and two innovative modules, a Adaptive Feature-refinement Unit (AFU) and a Double-head Gaze Regressor (DGR). The basic feature extractor captures the appearance feature of each facial image. The two proposed modules are designed to help extract informative and gaze-related appearance features. \looseness = -1 

\subsubsection{\textbf{Basic Appearance Feature Extractor}}
We employ the remarkable CNN-Transformer architecture~\cite{GazeTR} as the basic structure, where a ResNet is firstly adopted to acquire feature maps $f_i^{maps}\in \mathbb{R}^{W\times H\times C}$ of a given image $x_i \in X$ where $X = \{x_1, x_2, \dots,x_n\}$. Then those feature maps are reshaped into $W \times H$ patches $f_i^p\in  \mathbb{R}^{\left(W\times H\right)\times C}$,which are treated as a series of $C$-dimensional visual tokens. After adding an extra learnable token $f_i^{token}\in \mathbb{R}^{1\times C}$, which is used to aggregate the features of all the patches, we feed them into a Transformer with a learnable position embedding $f_i^{pos}\in \mathbb{R}^{\left(1+W\times H\right)\times C}$. Overall, we get the final primary appearance feature $f_i^{pry}\in \mathbb{R}^{1\times C}$ as Eq.~(\ref{eq:1}).
\begin{equation}
f_i^{pry}=\text{Transformer}\left(\left[{f_i^{token}} ; {f_i^p}\right]+{f_i^{pos}}\right)\left[0,:\right],
  \label{eq:1}
\end{equation}
where $[0,:]$ represents that the first row of the feature maps serves as the primary feature and $[ ; ]$ denotes the concatenation operation.\looseness = -1

\subsubsection{\textbf{Adaptive Feature-refinement Unit}}
 It is well-known that CLIP trained on large-scale image-text pairs from the Internet have achieved excellent performance in various face-related downstream tasks, including age estimation~\cite{TeachCLIPDevelopAge}, facial image editing~\cite{StyleCLIP}, etc~\cite{CLIPCluster}. Those remarkable applications demonstrate that the visual encoder of CLIP has powerful abilities to characterize prior appearance information of facial images. Undoubtedly, this appearance information is always mixed by crucial information that is beneficial for gaze estimation and redundant information which may harm the accuracy.  We hold an assumption that \textit{the components present in both the primary appearance feature and the prior appearance feature from CLIP are the most important ones for gaze estimation}. Therefore, the Adaptive Feature-refinement Unit (AFU) is designed to dynamically enhance the gaze-related contents contained in the primary appearance features with the guidance of the prior appearance features from CLIP.

Firstly, a group of gaze-aware attention units are introduced to calculate the attention maps of the prior appearance features  $f_i^{clip}$ from CLIP image encoder and raw feature maps $f_i^{maps}$ from Basic Appearance Feature Extractor, respectively (Eq.~(\ref{eq:9-2})).
% \begin{equation}
% f_i^{clip}= ImageEncoder(x_i),
% \label{eq:9-1}
% \end{equation}
% \begin{equation}
%     Q={fW}_Q, K=fW_k,
%     \label{eq:9-2}
% \end{equation}
% \begin{equation}
% M=\mathcal{G}\left(f\right)=Softmax\left(QK^T/\beta\right),
% \label{eq:9-3}
% \end{equation}
% \begin{equation}
% M_i^{clip}=\mathcal{G}\left(f_i^{clip}\right),M_i^{gaze}=\mathcal{G}\left(f_i^{maps}\right),
% \label{eq:9-4}
% \end{equation}
% \begin{equation}
%     M_i^{clip}=Softmax\left(f_i^{clip}{W_Q}(f_i^{clip}{W_K})^T/\beta\right)
% \label{eq:9-1}
% \end{equation}
\begin{equation}
\begin{aligned}
M_i^{clip}= \text{Softmax}\left({W_Q^{clip}}f_i^{clip}({W_K^{clip}}f_i^{clip})^T/\beta\right), \\
    M_i^{gaze}= \text{Softmax}\left({W_Q^{maps}}{f_i^{maps}}({W_K^{maps}}{f_i^{maps}})^T/\beta\right),
\end{aligned}
\label{eq:9-2}
\end{equation}
where $W_Q^{clip}$, $W_K^{clip}$, $W_Q^{maps}$, $W_K^{maps}$ are learnable
parameters of linear projections. 
The obtained attention maps $M$ capture the respective dependencies between components of the two features, which could reveal the valuable parts of them.

And then, to further activate the key gaze-related components of the primary appearance features, we integrate the two attention maps and refine the primary feature map to an enhanced one (Eq.~(\ref{eq:10})). Intuitively, the integrated attention map acts as a mask that highlights the important information involved in the primary appearance feature. \looseness = -1
\begin{equation}
{\hat{f}}_i^{maps}=(M_i^{clip}+M_i^{gaze})f_i^{maps}.
\label{eq:10}  
\end{equation}
Then we feed the enhanced feature maps ${\hat{f}}_i^{maps}$ into a Transformer to get an enhanced appearance feature ${{f}}_i^{img}$ following the process of the Basic Appearance Feature Extractor as Eq. ~(\ref{eq:1}).

By thoroughly exploring the relationships between components of primary appearance features and prior appearance features, the AFU could identify the vital information that is correlated to gaze directions and thus encourages the primary gaze estimation model to obtain informative and gaze-related appearance features. 

\subsubsection{\textbf{Double-head Gaze Regressor}}
Even with the enhanced features, the design of the regressor is also crucial for improving the model's generalization ability. Existing methods~\cite{FullFace,CA-Net} typically use a simple MLP-based architecture to regress features into gaze directions. As discussed in the early work~\cite{GPM}, the numerous parameters of MLP would easily overfit to gaze-irrelevant factors within the high-dimensional features during the mapping process. To mitigate the overfitting issues caused by MLP, we propose a Double-head Gaze Regressor (DGR). 
One of the regression head adopts the conventional MLP-based structure, which projects the enhanced gaze feature ${{f}}_i^{img}$ to a final gaze direction as Eq.~(\ref{eq:12}).
% \begin{equation}
% g_i=\text{MLP}({{f}}_i^{img}).
%     \label{eq:11}
% \end{equation}
Then we employ the $L_1$ loss (Eq. ~(\ref{eq:12})) to minimize the discrepancy between the estimated gaze direction $g_i$ and the ground truth $\hat{g_i}$,
\begin{equation}
g_i=\text{MLP}({{f}}_i^{img}), 
L_{gaze}= \frac{1}{N} \sum_{i=1}^{N} \left|\left|g_i-\hat{g}_i\right|\right|_1,
    \label{eq:12}
\end{equation}
where $N$ represents the number of image samples.

Inspired by the motivation of the dropout layers in neural networks~\cite{dropout}, we design a masked regression head. Specifically, we construct a mask $m_i\in \mathbb{R}^{1\times C}$ with the same size as $f_i^{img}$, whose elements are either 0 or 1. The numbers of 0s and 1s in the mask are manually adjusted by a drop ratio. For example, if the $f_i^{img}$ is a 32-dimensional vector and the drop ratio is set as $5/32$, the mask would include five 0s and twenty-seven 1s with random positions. Then we take the Hadamard product of the masks with our enhanced gaze features ${{f}}_i^{img}$ and get the masked features ${{f}}_i^{img\_m}$ (Eq.~(\ref{eq:13})).
\begin{equation}
{{f}}_i^{img\_m}=\ {{f}}_i^{img}\circ  M_i.
    \label{eq:13}
\end{equation}

Next, we utilize the sampling method of max-pooling to directly map the high-dimensional masked features ${{f}}_i^{img\_m}$ to 2D gaze vectors $\dot{g}_i$ without any parameters. Meanwhile, we also employ $L_1$ loss ($L_{mask}$ in Eq.~(\ref{eq:15})) to minimize the distance between the predicted gaze vector $\dot{g}$ and the ground truth $\hat{g}_i$.
% \begin{equation}
% \dot{g}_i=\ \text{MaxPooling}({{f}}_i^{img\_m}),
%     \label{eq:14}
% \end{equation}
\begin{equation}
\dot{g}_i=\ \text{MaxPooling}({{f}}_i^{img\_m}),  
L_{mask}\ =\frac{1}{N} \sum_{i=1}^{N} \left|\left|\dot{g}_i-\hat{g}_i\right|\right|_1.
    \label{eq:15}
\end{equation}
%where $N$ represents the size of batch.

The DGR guides the model to focus on all the dimensions of the features rather than overfit on several dimensions without increasing the number of parameters, which could promote the generalization ability of our gaze regressor.
\vspace{-0.2cm}
\subsection{Semantic Difference-aware Branch}
The Semantic Difference-aware branch is designed to implement the Differential Contrastive Training strategy. It aims to further enhance the extracted appearance features of the Visual Appearance-aware branch from the perspective of integrating gaze-related semantic information driven by the language-image alignment.

\subsubsection{\textbf{Differential Gaze Prompts}}
As indicated in Introduction, the challenge lies in the connections between the infinite continuous gaze direction of each facial image and the restricted language sentences. Intuitively, it is easier to describe the difference of gazes between two facial images than to give individual description of the gaze in each image. Therefore, we design a series of differential gaze prompts, each of which refers to a language sentence describing the correlations of gaze directions between a pair of images. 

To be specific, we pair each sample in a batch with every other one to form various image pairs and calculate their gaze differences with regards to their true gaze labels. 
The gaze difference is measured by the $L_1$ distances between the gaze directions of the image pair. 
Then, we categorize these image pairs into $K$ semantic similarity levels according to their gaze differences, and each level is attributed with a semantic grade name $t_i^{grade}$, e.g., `identical', `similar', `not similar'. 
Subsequently, a differential gaze prompt $T_i$ of each image pair is generated via combining a designed template\ $t_i^{template}$: `The directions of gaze in the two photos are \{\textit{grade name}\}.' with a corresponding semantic grade $t_i^{grade}$ (Eq.~(\ref{eq:6})). For instance, if $K$ is set to 2, all the selected image pairs are divided into `similar' and `not similar' groups. The image pairs with gaze differences that are from 0 to 0.2 are assigned into the `similar' grade, while image pairs with gaze differences that are over 0.2 are allocated to the `not similar' grade. The designed differential gaze prompts and their corresponding specific gaze differences are shown in Table ~\ref{tab:7}. And we evaluate the performance of different designs in Experiments. 
% Please add the following required packages to your document preamble:
% \usepackage{multirow}
\begin{table}[]
\small 
\caption{Design of the differential gaze prompts.}
\vspace{-0.2cm}
\label{tab:7}
\resizebox{0.9\linewidth}{!}{
\begin{tabular}{cccl}
\toprule
Template & $K$ & \begin{tabular}{c}
     Difference Interval
\end{tabular} & \begin{tabular}{c}
    Grade Names
\end{tabular} \\ \midrule
\multirow{10}{*}{\begin{tabular}[c]{@{}c@{}}`The directions \\ of gaze in the \\two photos are \\ \{Grade Name\}'.\end{tabular}} & \multirow{2}{*}{2} & {[}0, 0.2) & `similar' \\ 
 &  & {[}0.2, +$\infty$) & `not similar' \\ \cline{2-4}
 & \multirow{3}{*}{3} & {[}0,0.1) & `identical' \\ 
 &  & {[}0.1, 0.2) & `similar' \\ 
 &  & {[}0.2, +$\infty$) & `not similar' \\ \cline{2-4}
 & \multirow{5}{*}{5} & {[}0, 0.1) & `identical' \\
 &  & {[}0.1, 0.2) & `highly similar' \\
 &  & {[}0.2, 0.3) & `moderately similar' \\
 &  & {[}0.3, 0.5) & `slightly similar' \\
 &  & {[}0.5, +$\infty$) & `not similar' \\  \bottomrule
\end{tabular}
}
\end{table}

\subsubsection{\textbf{Differential Contrastive Training}} 
Given the differential gaze prompts $T_i$, a pre-trained CLIP text encoder is employed to encode it into a text embedding $(f_i^{text})$ as Eq.~(\ref{eq:6}),
\begin{equation}
T_i=[t_i^{template},t_i^{grade}], f_i^{text}=TextEncoder(T_i).
\label{eq:6}
\end{equation}

To further enhance the extracted appearance features, we innovatively realize a visual-semantic alignment between the CLIP text encoder and the Visual Appearance-aware branch. Ideally, the semantic enhanced visual features of facial images should be aligned with the above text embeddings. In other words, the compatibilities between visual embeddings of image pairs and the text embeddings of their corresponding differential gaze prompts should be higher than the compatibilities between image pairs and other misaligned descriptions. 
The visual embeddings of image pairs are derived from the concatenated features of the above visual branch of the selected two images, as Eq.~(\ref{eq:7}),
\begin{equation}
f_i^{pair} = \text{MLP}([f_{i1}^{img};f_{i2}^{img}]).
\label{eq:7}
\end{equation}
%where $f_{i*}^{img}$ represents the feature of the $i$-th image in the $j$-th image pair. 

A language-driven contrastive loss is thus designed as Eq.~(\ref{eq:8}),
\begin{equation}
\begin{split}
L_{align} = -\frac{1}{N_p}\sum_{i=1}^{N_p}log\frac{exp(f_i^{text}\cdot f_i^{pair}/\tau)}{\sum_{j =1}^{N_p}{exp(f_i^{text}\cdot f_j^{pair}/\tau)}}\\
-\frac{1}{N_p}\sum_{i=1}^{N_p}log\frac{exp(f_i^{pair}\cdot f_i^{text}/\tau)}{\sum_{j =1}^{N_p}{exp(f_i^{pair}\cdot f_j^{text}/\tau)}},
\end{split}
    \label{eq:8}
\end{equation}
where $N_p$ is the number of selected image pairs in one batch and $\tau$ is a temperature hyperparameter.
In our experiments, for each facial image in a sample batch, all the remaining samples in this batch are selected to constitute image pairs with it. Thus we can obtain $N_p = (N_b-1)\times N_b$ pairs in each sample batch and then partition them into $K$ semantic grades according to their true gaze labels. $N_b$ denotes the number of samples in a training batch. 

By minimizing the $L_{align}$, our Visual Appearance-aware branch could be endowed with the ability to perceive gaze semantic difference, thus takes full advantages of the gaze-related semantic information. The innovations could be illustrated as follows. 
On the one hand, rather than extracting features from individual images, by comparing different facial images, interactions of gaze between them will be characterized, which benefits the extraction of robust gaze-related features. 
On the other hand, the proposed Differential Contrastive Training strategy essentially simplifies the target words that the CLIP needs to understand by replacing unfamiliar phrases, e.g., `[0.54 (pitch), -0.20 (yaw)]' with familiar ones, e.g., `similar', `not similar' (Figure ~\ref{fig:1}). 
Moreover, the continuous gaze labels are transformed into discrete similarity levels of a pair of gazes. 
These could cleverly address the challenges that hinder the applications of CLIP in gaze estimation as discussed in Introduction. Eventually, by realizing the language-image alignment, the consistency between extracted features and gaze labels is maintained derivatively. Differential Contrastive Training helps to learn robust and pure gaze-related features.

\subsection{Optimization and Inference}
In training stage, our network is optimized by minimizing the total loss function as follows:
%\vspace{-0.2cm}
\begin{equation}
  L_{total}\ =\ L_{gaze}+ \alpha L_{mask}+\beta L_{align},  
  \label{eq:total}
 % \vspace{-0.2cm}
\end{equation}
where $\alpha$ and $\beta$ are hyperparameters to balance the losses. The text encoder of Semantic Difference-aware branch is frozen and the parameters of the primary gaze estimation network consisting of AFU and DGR should be updated. 

During the inference stage, the Semantic Differential-aware branch and the masked regression head of DGR should be cut off. In other words, we believe that with the Differential Contrastive Training, the primary gaze estimation model is capable of extracting informative gaze-related features.
Therefore, the facial images are fed into the Basic Appearance Feature Extractor and the AFU to obtain the enhanced appearance features. Finally, we employ the MLP-based head to project the enhanced features into final gaze directions. 
Moreover, some existing studies ~\cite{DifferentialGaze} have also employed differential ideas to remove gaze-irrelevant factors from extracted gaze features. However, they still need to implement the differential calculations during the inference stage, whose predicted results heavily rely on the selected reference samples. Superior to them, our proposed differential module is regarded as an assistant line, whose mission is to encourage the primary network to extract a robust feature that contains gaze-related information. Therefore, we achieve a simpler and more direct inference stage. 
\vspace{-10pt}
\section{Experiments}
\label{sec:exp}
\subsection{Datasets and settings}
Our method is evaluated on four popular gaze estimation datasets, which are MPIIFaceGaze ($\mathcal{D}_M$)~\cite{MPII}, EyeDiap ($\mathcal{D}_D$)~\cite{eye}, Gaze360 ($\mathcal{D}_G$)~\cite{Gaze360} and ETH-XGaze ($\mathcal{D}_E$)~\cite{eth} over within and cross-domain tasks. More details of datasets refer to supplementary materials.

For within-domain evaluation, the experiments are conducted on MPIIFaceGaze, EyeDiap and Gaze360 datasets. We perform leave-one-person-out evaluation on MPIIFaceGaze dataset and four-folder cross validation on EyeDiap dataset. As for the Gaze360 dataset, after removing the images without frontal faces, we select 84,902 images of 54 subjects for training and 16,000 images of 15 subjects for testing. For cross-domain evaluation, the Gaze360 and ETH-XGaze datasets are treated as source domains for training, while MPIIFaceGaze and EyeDiap are target ones for testing. Thus, we evaluate our method on four cross-domain tasks: $\mathcal{D}_E\rightarrow \mathcal{D}_M, \mathcal{D}_E\rightarrow \mathcal{D}_D, \mathcal{D}_G\rightarrow \mathcal{D}_M, \mathcal{D}_G\rightarrow \mathcal{D}_D$.
\vspace{-5pt}
\subsection{Implementation details}
We employ the pre-trained RN50 CLIP for the backbones of our Semantic Difference-aware branch and AFU, which consists of a ResNet50-based image encoder and a Transformer-based text encoder. During the training stage, the text encoder is frozen, while the image encoder would be fine-tuned.

In the within-domain evaluation, the Basic Appearance Feature Extractor is composed of a ResNet18 and a 6-layer Transformer. %Following~\cite{GazeTR}, we employ a model that is pre-trained on the ETH-XGaze dataset and would be fine-tuned on other datasets. 
Given the $224\times224$ input images, $7\times7\times32$ feature maps are generated from ResNet18 and then fed into a 6-layer Transformer with 8-head self-attention mechanism. Finally, we get a 32-dimensional image feature. In the cross-domain evaluation, the Transformer is replaced by a 3-layer MLP to mitigate the overfitting issues. 
The weighted factors $\alpha$ and $\beta$ in Eq.~(\ref{eq:total}) are set to (0.1, 0.1) for both within-domain tasks and cross-domain tasks.
\vspace{-5pt}
\subsection{Comparison with State-of-the-art Methods}
We compare our approach with the SOTAs (Table ~\ref{tab:2} and Table ~\ref{tab:3}). 
% The reported results come from~\cite{GFAL} or their original papers. 
% \textbf{DCGaze} denotes our full model. \textbf{DCGaze$^*$} denotes the sub-model that removes the AFU.
The reported results come from~\cite{GFAL} or their original papers. We announce two versions of our method.
\textbf{DCGaze-AFU} denotes our full model. \textbf{DCGaze-Base} denotes the model without AFU.

% In addition to the complete model, we also evaluates the performance of the degrade model  DCGaze$^-$ which is the degrade network without the AFU. 

\textbf{Within-domain.}
%We roughly classify the compared methods into CNN-based methods ( Itracker~\cite{Itracker}, FullFace~\cite{FullFace}, RT-Gene~\cite{Rt}, Dilated-Net~\cite{Dilated-Net}, Gaze360~\cite{Gaze360}, CA-Net~\cite{CA-Net}), FAR-Net~\cite{FAR} and transformer-based methods (oh et al.~\cite{oh}, GazeTR~\cite{GazeTR}, SUGE~\cite{SUGE}). 
We roughly classify the compared methods into CNN-based methods (upper part of Table ~\ref{tab:2}) and Transformer-based methods (bottom part of Table ~\ref{tab:2}). 
% As the results shown in Table ~\ref{tab:2}, our approach outperforms all within-domain methods on the MPIIFaceGaze and EyeDiap datasets. It also achieves performance comparable to the SUGE~\cite{SUGE} on Gaze360 dataset. The results prove that our DCGaze can strengthen the gaze-related appearance and semantic information in gaze features, leading to higher accuracy.
Our DCGaze-AFU outperforms all within-domain methods on the MPIIFaceGaze and EyeDiap datasets. It also achieves performance comparable to the SUGE~\cite{SUGE} on Gaze360 dataset. The results prove that our approach can strengthen the gaze-related appearance and semantic information in gaze features, leading to higher accuracy.

\textbf{Cross-domain.}
To further demonstrate the generalizability of our method, we conduct experiments on unsupervised domain adaptation (UDA) tasks and supervised domain adaptation (SDA) tasks. 
% In UDA settings, the models are trained on source domain while testing on unseen target domain. $|\mathcal{D}_t|$ samples from target domain could be randomly selected for further adaptation (fine-tuning or co-training). 
In all settings, the models are trained on source domain while tested on unseen target domain. $|\mathcal{D}_t|$ samples from target domain whose labels are unseen in UDA tasks while avaliable in SDA tasks could be randomly selected for further adaptation. 
% As the results shown in the Table ~\ref{tab:3}, our sub-model DCGaze$^*$  achieves the best performance on $\mathcal{D}_E\rightarrow \mathcal{D}_M$, $\mathcal{D}_G\rightarrow \mathcal{D}_M$ and $\mathcal{D}_G\rightarrow \mathcal{D}_D$ tasks. 
% As the results shown in the Table ~\ref{tab:3}, our sub-model DCGaze$^*$  achieves the best performance on three tasks both in UDA and SDA tasks. 
As the results shown in the Table ~\ref{tab:3}, our  DCGaze-Base achieves the best performance on three tasks both in UDA and SDA tasks. 
Those results also demonstrate the effectiveness of our proposed DCTrain strategy.\looseness=-1

% However, the complete DCGaze demonstrates performance inferior to the degrade network DCGaze$^*$. We analyze the possible reasons. 
% CLIP is likely to encode scene contexts (e.g., image resolution, illumination, etc.) into its obtained appearance embeddings. For cross-domain tasks, the test samples own significantly different environments with training ones. Thus the AFU is more adept at capturing the scene information of the source domain, which may disturb the estimation of disjoint target samples. 
% Nevertheless, those results demonstrates the effectiveness of our proposed Differential Contrastive Training strategy. 

% Since CLIP has strong capability in extracting appearance information, it can capture gaze-related factors in specific scene contexts (e.g., image resolution, illumination, etc.) and dynamically inject them into the original features through the AFU. However, different datasets are collected in different scenes. For example, the Gaze360 dataset is collected in surveillance settings, while the EyeDiap dataset is collected in laboratory environments. This difference leads to the AFU being more adept at capturing appearance representations from the scene of the source domain, but it struggles to adapt to the target domain during cross-domain tasks, resulting in a decline in performance.
\begin{table}[t]
\small
    \caption{Performance on within-domain tasks.   Results reported are angular errors in degrees. \textbf{Bold} and \underline{underline} indicate the best and the second best results.}
  \label{tab:2}
  \vspace{-8pt}
    \centering
    \begin{tabular*}{\linewidth}{@{}@{\extracolsep{\fill}}lcccc@{}}
     \toprule
    Method & Refs & $\mathcal{D}_M$ & $\mathcal{D}_D$ & $\mathcal{D}_G$ \\
    \midrule
        Itracker~\cite{Itracker} & 2016 CVPR & 6.20 & 9.93 & - \\
        FullFace~\cite{FullFace} & 2017 CVPRW & 4.93 & 6.53 & 14.99 \\
        RT-Gene~\cite{Rt} & 2018 ECCV & 4.30 & 5.90 & - \\
        Dilated-Net~\cite{Dilated-Net} & 2018 ACCV & 4.42 & 6.19 & 13.73 \\ 
        Gaze360~\cite{Gaze360}  & 2019 ICCV & 4.06 & 5.36 & 11.04 \\ 
        FAR-Net~\cite{FAR} & 2020 TIP & 4.30 & 5.71 & - \\
        CA-Net~\cite{CA-Net}  & 2020 AAAI & 4.27 & 5.27 & 11.20 \\ 
     \midrule
      GazeTR~\cite{GazeTR} & 2021 ICPR & 4.00 & 5.17 & 10.62 \\ 
        Oh et al. ~\cite{oh} & 2022 CVPRW & 4.04 & 5.25 & 10.70 \\       
        SUGE~\cite{SUGE}  & 2024 AAAI & 4.01 & \underline{5.04} & \textbf{10.51} \\ 
    \midrule
        \textbf{%DCGaze$^*$
         DCGaze-Base} & \textbf{Ours} & \underline{3.76} & 5.15 & 10.58 \\ 
        \textbf{%DCGaze
         DCGaze-AFU} & \textbf{Ours} & \textbf{3.71} & \textbf{4.97} & \underline{10.54} \\ 
         \bottomrule
    \end{tabular*}
\end{table}

\begin{table}[t]
    \centering
    \small
    \caption{Performance on cross-domain tasks.   $|\mathcal{D}_t|$: the number of samples for fine-tuning or co-training. \textbf{Bold} and \underline{underline} indicate the best and the second best results.}
    \label{tab:3}
    \begin{tabular*}{\linewidth}{@{}@{\extracolsep{\fill}}l@{} c@{}  c@{} c@{} c@{} c@{} c@{}c}
     \toprule
     Tasks & Methods & Refs &  $|\mathcal{D}_t|$ &
     \begin{tabular}{c@{}} $\mathcal{D}_E$ \\ $\rightarrow \!\mathcal{D}_M$ \end{tabular} & \begin{tabular}{c@{}} $\mathcal{D}_E$ \\ $\rightarrow \!\mathcal{D}_D$ \end{tabular} & \begin{tabular}{c@{}} $\mathcal{D}_G$ \\ $\rightarrow \!\mathcal{D}_M$ \end{tabular} & \begin{tabular}{c@{}} $\mathcal{D}_G$ \\ $\rightarrow \!\mathcal{D}_D$ \end{tabular} \\ 
    \midrule   
    \multirow{12}{*}{UDA} & ADDA~\cite{ADDA} & 2017 CVPR & 500 & 6.65 & 8.24 & 6.27 & 9.53 \\ 
    &GazeAdv~\cite{GazeAdv} & 2019 CVPR& 100 & 6.36 & 7.62 & 7.54 & 8.43 \\ 
    &Gaze360~\cite{Gaze360} & 2019 ICCV & 100 & 6.24 & 7.47 & 7.17 & 7.66 \\ 
    &DAGEN~\cite{DAGEN} & 2020 ACCV  &500 & 5.73 & 6.77 & 7.38 & 8.00 \\ 
    &UMA ~\cite{UMA} & 2020 CVPR &100 & 7.52 & 12.37 & 8.51 & 19.32 \\
    &PnP-GA~\cite{Pnp-GA} & 2021 ICCV& 10 & 5.53 & 5.87 & 6.18 & 7.92 \\ 
    &CRGA ~\cite{CRGA} & 2022 CVPR &>0 & 5.48 & \underline{5.66} & 5.89 & \underline{6.49} \\
    &RUDA~\cite{RUDA} & 2022 CVPR & 100 & 5.70 & 7.52 & 6.20 & 7.02 \\ 
    &DCUA~\cite{DCUA} & 2024 TMM & 100 & 7.31 & 5.95 & \underline{5.59} & \textbf{6.40} \\
    &PnP-GA$^+$ ~\cite{pnp+} & 2024 TPAMI & 10 & \underline{5.34} & 5.73 & 6.10 & 7.62 \\
    &\textbf{%DCGaze$^*$
     DCGaze-Base} &\textbf{Ours} &\centering 100 & \textbf{5.26} & \textbf{5.65} & \textbf{5.53} & 6.97 \\ 
    &\textbf{%DCGaze
     DCGaze-AFU} & \textbf{Ours} &\centering 100 & 6.47 & 6.70 & 6.45 & 7.74 \\ 
        \midrule
    \multirow{3}{*}{SDA} &CLIP-Gaze~\cite{CLIPGaze} & 2024 AAAI& 100 & \underline{4.45} &\textbf{5.27} & \underline{4.94} & \underline{5.60} \\
    &\textbf{%DCGaze$^*$
     DCGaze-Base} &\textbf{Ours} &\centering 100 & \textbf{4.37} & \underline{5.36} & \textbf{4.83} & \textbf{5.49} \\ 
    &\textbf{%DCGaze
     DCGaze-AFU} & \textbf{Ours} &\centering 100 & 6.00 & 6.01 & 5.25 & 6.21 \\
         \bottomrule
    \end{tabular*}
\end{table}

\subsection{Ablation Study}
\subsubsection{\textbf{Effectiveness of the DCTrain strategy.}} 
We treat the Semantic Difference-aware branch along with our Differential Contrastive Training strategy as a plug-and-play module, and evaluate its properties via integrating it into various basic gaze estimation models. 
% brings performance improvements to various basic gaze estimation models.
% ResNet18~\cite{Resnet}, ResNet50~\cite{Resnet}, GazeTR~\cite{GazeTR} and evaluate them on both within-domain and cross-domain tasks, 
The results are shown in Table ~\ref{tab:9} and Table ~\ref{tab:10}. 
% The integration of the Semantic Difference-aware branch improves the performance of the primary gaze estimation network, which indicates that the Differential Contrastive Training strategy we proposed can indeed help the model extract robust features.
our DCTrain strategy could bring performance improvements to those various basic models. It demonstrates that DCTrain does develop the potentials of the pre-trained CLIP in boosting gaze estimation performance.
\begin{table}[h]
\small
\caption{Effectiveness of the DCTrain on within-domain tasks.}
\vspace{-5pt}
\label{tab:9}
\resizebox{0.9\linewidth}{!}{
\begin{tabular*}{\linewidth}{@{}@{\extracolsep{\fill}}c c c @{} c}
\toprule
Model & MPII & EyeDiap & Gaze360 \\
\midrule
Resnet18 & 4.15 & 5.67 & 12.40 \\
Resnet18+DCTrain & 4.15 & \textbf{5.63} & \textbf{12.09} \\
\midrule
Resnet50 & 4.30 & 5.67 & 11.77 \\
Resnet50+DCTrain & \textbf{4.19} & \textbf{5.59} & \textbf{11.68} \\
\midrule
GazeTR & 4.13 & 5.23 & 10.76 \\
GazeTR+DCTrain & \textbf{3.73} & \textbf{5.15} & \textbf{10.56}\\
\bottomrule
\end{tabular*}
}
\vspace{-5pt}
\end{table}

\begin{table}[h]
\small
\caption{Effectiveness of the Differential Contrastive Training on cross-domain tasks (SDA).}
\label{tab:10}
\vspace{-5pt}
\resizebox{0.9\linewidth}{!}{
\begin{tabular*}{\linewidth}{@{}@{\extracolsep{\fill}}c c c c c}
\toprule
Model & \begin{tabular}{c@{}} $\mathcal{D}_E$ \\ $\rightarrow \mathcal{D}_M$ \end{tabular} & \begin{tabular}{c@{}} $\mathcal{D}_E$ \\ $\rightarrow \mathcal{D}_D$ \end{tabular} & \begin{tabular}{c@{}} $\mathcal{D}_G$ \\ $\rightarrow \mathcal{D}_M$ \end{tabular} & \begin{tabular}{c@{}} $\mathcal{D}_G$ \\ $\rightarrow \mathcal{D}_D$ \end{tabular} \\
\midrule
Resnet18 & 5.56 & 6.83 & 8.81 & 8.92\\
Resnet18+DCTrain & \textbf{5.55} & \textbf{6.41} & \textbf{8.02} & \textbf{8.60} \\
\midrule
Resnet50 & 5.25 & 6.51 & 4.93 & 5.41 \\
Resnet50+DCTrain & \textbf{5.21} & \textbf{6.33} & \textbf{4.87} & \textbf{5.36} \\
\bottomrule
\end{tabular*}
}
\vspace{-5pt}
\end{table}

\subsubsection{\textbf{Ablation Study of Proposed Modules}}
% To investigate the effectiveness of each proposed module, we compare the performance of some degraded models on within-domain tasks. Based on the fully model, we remove one proposed module each time and get the results shown in ~\ref{tab:4}. ‘Baseline’ refers the backbone model that consists of a CNN-Transformer-based Feature Extractor and an MLP-based regressor. No matter which module is invalidated, the performance decreases. This proves the importance of each component. 1) By comparing `w/o LDM' with `Full model', it demonstrates that the LDM can help Basic Appearance Feature Extractor to capture robust and pure gaze-related features via image-text alignment. 2) By comparing `w/o AFU' with `Full model', it illustrates that the AFU can further enhance valuable partitions of the gaze features through integrating generalized embeddings of CLIP. 3) By comparing `w/o DGR' with `Full model', the Double-head Gaze Regressor have been proven effective in reducing degrees of freedom and improving generalizability.
To investigate the effectiveness of each proposed component, we compare the performance of some degraded models on both within and cross-domain tasks. Based on the fully model, we remove one proposed component each time and get the results in Table ~\ref{tab:4} and Table ~\ref{tab:8}. The $1st.$ rows refer the baseline models. In within-domain evaluation, the baseline model consists of a CNN-Transformer-based Feature Extractor and an MLP-based regressor. In cross-domain evaluation, the Transformer part of the Feature Extractor is replaced by a 3-layer MLP. In within-domain tasks, no matter which component is invalidated, the performance decreases. This proves the importance of each of them. 
1) By comparing the $2nd.$ rows with the $5th.$ rows (full model), it demonstrates that the Semantic Difference-aware branch can help the visual branch capture robust and pure gaze-related features via image-text alignment. 
2) By comparing the $4th.$ rows with the $5th.$ rows, the DGR has been proven effective in reducing degrees of freedom and improving generalizability.
3) By comparing the $3rd.$ rows with the $5th.$ rows, AFU could bring improvements to within-domain tasks but lead a decline in performance for cross-domain ones. 
We analyze the possible reasons. 
CLIP is likely to encode scene contexts (e.g., image resolution, illumination, etc.) into its obtained appearance embeddings. For cross-domain tasks, the test samples own significantly different environments with training ones. Thus the AFU is more adept at capturing the scene information of the source domain, which may disturb the estimation of disjoint target samples.

% it illustrates that, on within-domain tasks, the AFU can further enhance valuable partitions of the appearance features with the guidance of the prior appearance features of CLIP. Howerever, on cross-domain tasks, the AFU leads to a decline in performance. 

% However, the complete DCGaze demonstrates performance inferior to the degrade network DCGaze$^*$. 

\begin{table}[t]
  \centering
  \small
  % \begin{tabular}{l c c c@{}}
  %   \toprule
  %   Method & MPII & EyeDiap & Gaze360 \\ 
  %   \midrule 
  %       Baseline & 4.13 & 5.23 & 10.76 \\ 
  %       w/o LDM & 3.77 & 5.06 & 10.65 \\ 
  %       w/o AFU & 3.76 & 5.15 & 10.58 \\ 
  %       w/o DGR & 3.77 & 5.18 & 10.58 \\ 
  %       \textbf{Full model} & \textbf{3.71}& \textbf{4.97} & \textbf{10.54} \\ 
  %   \bottomrule
  % \end{tabular}
\vspace{-5pt}
\caption{Ablation study results of proposed components on within-domain tasks.}
  \label{tab:4}
  \vspace{-8pt}
  \resizebox{0.9\linewidth}{!}{
  \begin{tabular*}{\linewidth}{@{}@{\extracolsep{\fill}}c c c | c c c}
    \toprule
    DCTrain &AFU &DGR & MPII & EyeDiap & Gaze360 \\ 
    \midrule 
        - & - & - & 4.13 & 5.23 & 10.76 \\ 
        - & \checkmark & \checkmark & 3.77 & 5.06 & 10.65 \\ 
        \checkmark &- &\checkmark & 3.76 & 5.15 & 10.58 \\ 
        \checkmark&\checkmark&- & 3.77 & 5.18 & 10.58 \\ 
        \checkmark & \checkmark & \checkmark & \textbf{3.71}& \textbf{4.97} & \textbf{10.54} \\ 
         \bottomrule
  \end{tabular*}
  }
\vspace{-8pt}
\end{table}

% \begin{table}[t]
%     \centering
%     \small
%     \caption{Ablation study results of proposed components on cross-domain tasks.}
%     \label{tab:8}
%     \begin{tabular*}{\linewidth}{@{}ccc|cccc@{}}
%      \toprule
%     DCTrain &AFU &DGR & \begin{tabular}{@{}c@{}} $\mathcal{D}_E$ \\ $\rightarrow \mathcal{D}_M$ \end{tabular} & \begin{tabular}{@{}c@{}} $\mathcal{D}_E$ \\ $\rightarrow \mathcal{D}_D$ \end{tabular} & \begin{tabular}{@{}c@{}} $\mathcal{D}_G$ \\ $\rightarrow \mathcal{D}_M$ \end{tabular} & \begin{tabular}{@{}c@{}} $\mathcal{D}_G$ \\ $\rightarrow \mathcal{D}_D$ \end{tabular}\\ 
%     \midrule 
%         - & - & - & 7.14 & 7.83 & 7.10 & 8.59 \\ 
%         - & \checkmark & \checkmark & 6.47 & 6.20 & 5.65 & 6.60 \\ 
%         \checkmark &- &\checkmark & \textbf{4.37} & \textbf{5.36} & \textbf{4.83} & \textbf{5.49} \\ 
%         \checkmark & \checkmark & - & 6.48 & 6.75 & 5.97 &6.89 \\ 
%          \checkmark & \checkmark & \checkmark & 6.21 & 6.05 & 5.63 & 6.33 \\ 
%          \bottomrule
%     \end{tabular*}

% \end{table}
\begin{table}[t]
    \centering
    \small
    \caption{Ablation study results of proposed components on cross-domain tasks (SDA).}
    \label{tab:8}
    \vspace{-8pt}
    \resizebox{0.9\linewidth}{!}{
    \begin{tabular*}{\linewidth}{@{\extracolsep{\fill}}ccc|cccc@{}}
     \toprule
    DCTrain &AFU &DGR & \begin{tabular}{@{}c@{}} $\mathcal{D}_E$ \\ $\rightarrow \mathcal{D}_M$ \end{tabular} & 
    \begin{tabular}{@{}c@{}} $\mathcal{D}_E$ \\ $\rightarrow \mathcal{D}_D$ \end{tabular} & 
    \begin{tabular}{@{}c@{}} $\mathcal{D}_G$ \\ $\rightarrow \mathcal{D}_M$ \end{tabular} & 
    \begin{tabular}{@{}c@{}} $\mathcal{D}_G$ \\ $\rightarrow \mathcal{D}_D$ \end{tabular} \\ 
    \midrule 
        - & - & - & 7.14 & 7.83 & 7.10 & 8.59 \\ 
        - & \checkmark & \checkmark & 6.47 & 6.20 & 5.65 & 6.60 \\ 
        \checkmark &- &\checkmark & \textbf{4.37} & \textbf{5.36} & \textbf{4.83} & \textbf{5.49} \\ 
        \checkmark & \checkmark & - & 6.48 & 6.75 & 5.97 &6.89 \\ 
        \checkmark & \checkmark & \checkmark & 6.21 & 6.05 & 5.63 & 6.33 \\ 
     \bottomrule
    \end{tabular*}
    }
    \vspace{-8pt}
\end{table}

\begin{table*}[h]
  \centering
  \small
    \caption{Analysis of differential gaze prompts.  The fixed template is `The directions of gaze in the two photos are \{\textit{Grade Name}\}.' }
  \label{tab:5}
  \vspace{-8pt}
  \resizebox{0.9\textwidth}{!}{
  \begin{tabular*}{\linewidth}{@{}@{\extracolsep{\fill}}c c c c c c}
    \toprule
    $K$ & Template & Grade Names & MPII & EyeDiap & Gaze360 \\ 
    \midrule 
        2 & & `similar', `not similar'& 3.73 & 5.08 & 10.56  \\
        3 & {\centering fixed} & `identical', `similar', `not similar' & 3.82 & 5.13 & 10.60  \\
        5 & & `almost identical', `extremely similar', `similar', `a little similar', `different' & 3.75  & 5.02 & 10.58  \\
        \midrule 
        5 & fixed &`identical', `highly similar', `moderately similar', `slightly similar', `not similar' & \textbf{3.71} & \textbf{4.97} & 10.54  \\
        \midrule 
        5 & learnable & `identical', `highly similar', `moderately similar', `slightly similar', `not similar' & 3.81 & 5.09 & \textbf{10.53}  \\ 
    \bottomrule
  \end{tabular*}
  }
  \vspace{-8pt}
\end{table*}

\subsubsection{\textbf{Analysis of Differential Gaze Prompts}}
In this section, we evaluate our DCGaze under different prompt designs (Table ~\ref{tab:5}). Firstly, we vary the number of grade prompts $K$ from 2 to 5. The results show that increasing the number of grades can better help the model to identify the subtle gaze differences between facial images, leading to better feature representations. However, the more levels there are, the more difficult manual designs of textual prompts become. Thus, we select 5 grades in our experiments. Secondly, we explore different strategies to describe these prompts. To be specific, we use different words for each differential grade name or propose a learnable prompt template following the CoOp method~\cite{COOP}. Two main observations could be concluded from the comparisons. 1) The manual prompts combining with degree adverbs can convey more precise semantic information, which lead to better results. 2) The learnable prompt template performs worse than the fixed one, whose possible reason is that the fixed prompt template could clearly describe the gaze difference between images while the learnable one may introduce some noise during the learning process.

\subsubsection{\textbf{Analysis of Feature Refinement Strategy}}

As discussed in Section 3.1.2, a novel Adaptive Feature-refinement Unit is proposed to dynamically enhance the gaze-related contents of primary appearance features. In order to evaluate the properties of this unit, we compare it with several commonly used feature fusion methods on within-domain tasks. 

    $\bullet$ `Concatenation’: We concatenate the prior appearance feature of CLIP and the primary appearance feature, and feed them into a fully connected layer to preserve the feature dimensions.
  
    $\bullet$ `Cross-Attention’: We treat the primary appearance feature $f_{pry}$ as query $Q$ and the prior appearance feature $f_{clip}$ as key $K$ and value $V$. The final fused features $f^{img}$ is computed as Eq.~(\ref{eq:18}):
     \begin{equation}
     \centering
     \begin{split}
     Q_{pry}=f^{pry}W_Q, 
     K_{clip}=&f^{clip}W_K, 
 {V}_{clip}=f^{clip}W_V,\\
     f^{img}=\text{Softmax}&(Q_{pry}K_{clip}^T/\beta){V}_{clip}.
     \end{split}
     \label{eq:18}
     \end{equation}
     
    $\bullet$ `Gated Information Fusion’: Following~\cite{Gate}, the prior appearance feature $f^{clip}$ is processed by a sigmoid operation to generate a mask, which is used to activate the primary appearance feature $f^{img}$ via Hadamard product. The detailed process refers to Eq.~(\ref{eq:19}):
     \begin{equation}
    f^{img}=\ f^{pry}\circ \text{Sigmoid}(f^{clip}), 
     \label{eq:19}
     \end{equation}

As shown in Table ~\ref{tab:6}, our AFU consistently outperforms all compared methods, which indicates that it could effectively highlight vital appearance information to enhance the primary gaze feature.
\begin{table}[h]
  \centering
  \small
    \caption{Analysis of feature fusion strategy.}
  \label{tab:6}
  \vspace{-8pt}
  \resizebox{0.9\linewidth}{!}{
  \begin{tabular*}{\linewidth}{@{}@{\extracolsep{\fill}}c c c c }
    \toprule
   Fusion methods & MPII & EyeDiap & Gaze360 \\  
    \midrule 
        Concatenation & 3.77 & 5.29 & 10.57 \\ 
        Cross-Attention & 3.80 & 5.21 & 10.56 \\ 
        Gated Information Fusion & 3.78 & 5.13 & 10.59 \\ 
       \textbf{ AFU (proposed)} & \textbf{3.71} & \textbf{4.79} & \textbf{10.54} \\ 
    \bottomrule
  \end{tabular*}
  }
\end{table}

% \subsubsection{Ablation Study of image pairs Selection Strategy}
% The performance of the Differential Contrastive Training strategy can be influenced by the selection of image pairs. In our approach, we select all distinct images in a batch to form $(N-1)\times(N-1)$ image- pairs, which we believe effectively explores the relationships between samples within a batch. 

% To evaluate the effectiveness of this strategy, we propose a MemoryBank strategy as an alternative sample selection strategy. To be specific, we construct a fixed-size memory bank whose size is $M$ based on a queue to store samples. When the queue is full, we discard the earliest entered samples and replace them with new ones. During the image-pair construction, we form $N\times M$ sample pairs by combining the current batch's samples with all the samples in the memory bank. These image pairs are then used for Differential Contrastive Training strategy.

% As the results shown in Table ~\ref{tab:11}, our method outperforms all the MemoryBank with different size. It proves that an excess of sample pairs can negatively impact the performance of Differential Contrastive Training Strategy.

% \begin{table}[h]
% \label{tab:11}
% \caption{Ablation Study of image pairs Selection Strategy}
% \begin{tabular*}{\linewidth}{@{}@{\extracolsep{\fill}}ccccc}
% \toprule
% Selection Strategy & Size & MPII & EyeDiap & Gaze360 \\
% \midrule
% \multirow{2}{*}{MemoryBank} & $N\times2$ &  & 5.24 & 10.54 \\
%  & $N\times4$ &  & 5.29 & 10.57 \\
% \midrule
% Ours & / & 3.73 & 5.15 & 10.50\\
% \bottomrule
% \end{tabular*}
% \end{table}
\vspace{-10pt}
\subsection{More Discussion \protect \footnote{In supplementary materials, we conduct more experiments to evaluate the properties of our proposed modules.}}
\subsubsection{\textbf{
Visualization of Obtained Gaze Features}}
\label{vis}
To quantitatively demonstrate the advantages of our enhanced gaze features, we visualize the distribution of the training samples of Gaze360 and ETH-XGaze datasets by t-SNE~\cite{sne}, following~\cite{ContrastiveRegressionDomain}. 
We select all the training samples of Gaze360 dataset and randomly select 10000 training samples from ETH-XGaze dataset. 
In the scatter plot, the samples are clustered by KMeans~\cite{kmeans} according to their gaze labels, so that the samples with similar gaze directions share similar colors.

\begin{figure}[h]
  \centering
  %\fbox{\rule{0pt}{2in} \rule{0.9\linewidth}{0pt}}
   \includegraphics[width=0.8\linewidth]{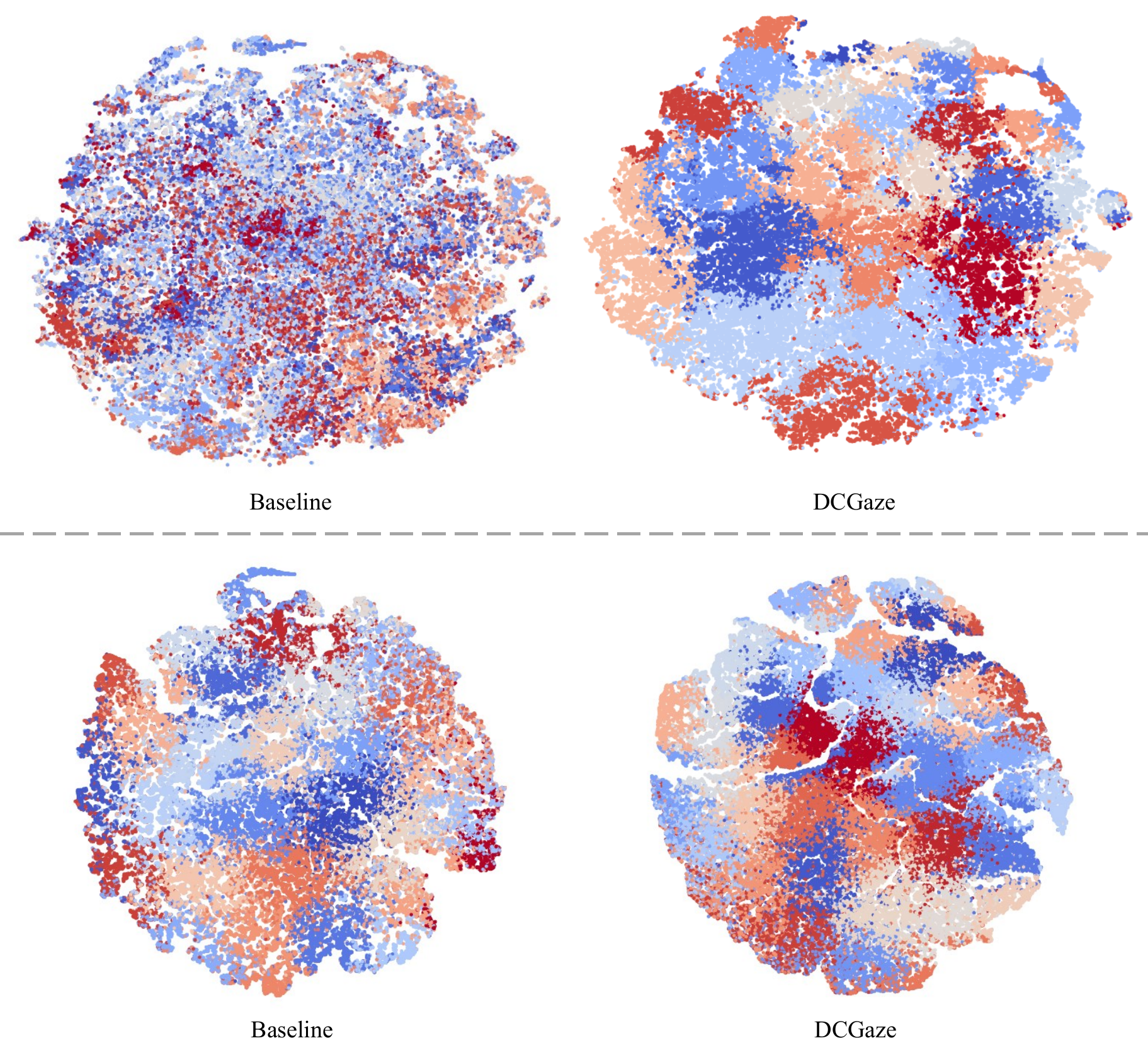}
   \caption{Visualization of obtained gaze features of Gaze360 (upper) and ETH-XGaze (bottom) datasets.}
   \label{fig:4}
\end{figure}
The feature distributions of the Baseline and our DCGaze are shown in Figure~\ref{fig:4}. 
As shown in the left figure, the sample points of baseline are scattered in a chaotic manner. By contrast, feature points of our enhanced features are distributed in an organized way, in which the features with similar gaze directions are clustered and can be regressed to similar gazes. It demonstrates our enhanced features being purified gaze-related ones, which effectively improve the discriminability and generalization.

\subsubsection{\textbf{Visualization of Estimated Gazes}}
\begin{figure}[h]
  \centering
  %\fbox{\rule{0pt}{2in} \rule{0.9\linewidth}{0pt}}
   \includegraphics[width=0.9\linewidth]{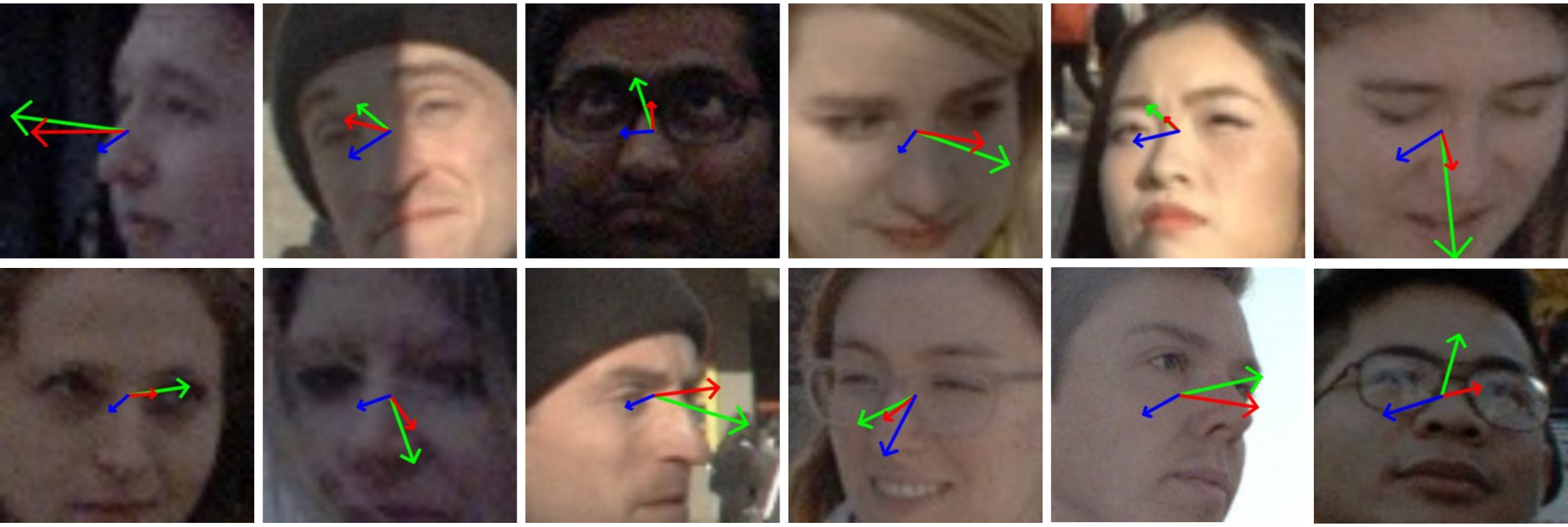}
   \caption{Visualization of estimated gazes.}
   \label{fig:5}
\end{figure}

 We further visualize the true gazes (green arrow), the predicted gaze directions of baseline (blue arrow) and the predicted gaze directions of DCGaze (red arrow). We select several test samples of Gaze360 dataset in different conditions including extreme head poses, dark illumination and low quality. The visualized results are shown in Figure~\ref{fig:5}. Compared to baseline method, the estimated results of our method are closer to ground truths in most conditions. 

\section{Conclusion}
\label{sec:con}
 \frenchspacing
In this paper, we have proposed a novel Differential Contrastive Gaze Estimation network (DCGaze) for gaze estimation, which leverages the powerful capabilities of the pre-trained CLIP to improve the representation capacity of a primary network by a novel Differential Contrastive Training strategy. 
Firstly, in the Visual Appearance-aware branch, the AFU and DGR have been proposed to help the primary gaze estimation model to capture informative and gaze-related features. Besides, to realize the Differential Contrastive Training, the Semantic Difference-aware branch has constructed image pairs and designed differential gaze prompt for each of them. Through visual-semantic alignment, the primary gaze estimation network has been endowed with the ability of extracting robust and pure gaze-related features. In extensive experiments, DCGaze has achieved remarkable performance on both within-domain tasks and cross-domain tasks. 
\textit{Limitation:} DCGaze is not a lightweight model, which would limit its deployment on edge devices in real application scenarios. Therefore, in the future, we would keep improving DCGaze in terms of reducing its scalability while maintaining the precision of prediction. \protect \footnote{In supplementary materials, we discuss the scalability of DCGaze compared to other methods.}

%%
%% The acknowledgments section is defined using the "acks" environment
%% (and NOT an unnumbered section). This ensures the proper
%% identification of the section in the article metadata, and the
%% consistent spelling of the heading.
\begin{acks}
This work is supported by Beijing Natural Science Foundation L251032, 4242028, the National Natural Science Foundation of China 62376021, Natural Science Foundation of Hebei Province F2025105036.
\end{acks}

%%
%% The next two lines define the bibliography style to be used, and
%% the bibliography file.
\bibliographystyle{ACM-Reference-Format}
\balance
\bibliography{sample-base}

%%
%% If your work has an appendix, this is the place to put it.
\renewcommand{\thefigure}{S-\arabic{figure}}
\setcounter{equation}{0}
\renewcommand{\theequation}{S-\arabic{equation}}
\setcounter{table}{0}
\renewcommand{\thetable}{S-\arabic{table}}
\appendix
\section{Details of the early attempt in Introduction}
\label{sec:earlyattempt}

\begin{figure}[h]
    \centering
\includegraphics[width=\linewidth]{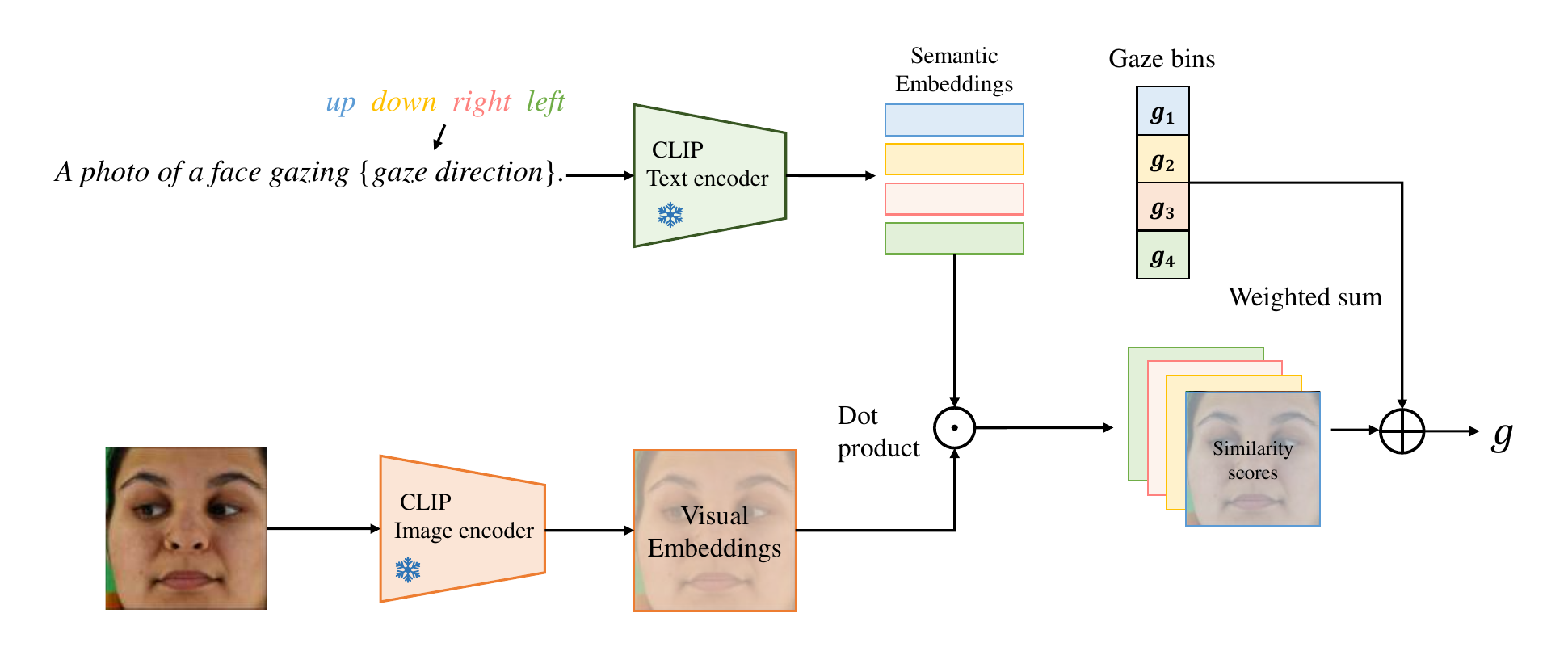}
    \caption{ The pipeline of our early attempt.}
    \label{fig:s1}
\end{figure}

As discussed in Introduction, to explore this concern `\textbf{\textit{can CLIP understand human gaze?}}’, we make a simple early attempt. The pipeline of our conducted experiment is shown in Figure~\ref{fig:s1}. 

Firstly, we define four gaze bins that align with the gaze directions of  [0,$\frac{\pi}{2}$], [0,-$\frac{\pi}{2}$], [$\frac{\pi}{2}$,0] and [-$\frac{\pi}{2}$,0], respectively. Those gaze bins are treated as gaze prototypes. Then, a series of semantic descriptions are designated to describe those prototypes, including [`up’, `down’, `left’, `right’]. For example, ‘up’ means that the person in the facial image who is looking up aligns with the gaze bin whose pitch is $\frac{\pi}{2}$ and yaw is 0. Subsequently, a manual text prompt in the form of `A photo of a face gazing \{\textit{gaze direction}\}.’ is designed for each gaze bin. The correspondences among prototypes, gaze bins and prompts are shown in Table~\ref{tab:s1}.
For each test facial image, we firstly feed it into the image encoder of the pre-trained CLIP (RN50-based version) and obtain a $C$-dimensional visual embedding $ f_i^{clip}$ as Eq.~\ref{eq:s1}. Meanwhile, the text prompts of the above four gaze prototypes are injected into the text encoder of the pre-trained CLIP and we can get four semantic embeddings as Eq.~\ref{eq:s2}. Then we calculate the cosine similarity between the visual embedding ($f_i^{clip}$)  and each of the semantic embedding ($f_j^{text}$) as Eq.~\ref{eq:s3}. The similarity score $S_{ij}$ indicates how similar the test facial image ($x_i$) is to the text prompt of $j-th$ gaze prototype. To get the final gaze direction $g_i$ of the test image, we linearly combine the gaze bins $g_j^{bin}$ based on calculated similarity scores as Eq.~\ref{eq:s4}.
\begin{equation}
f_i^{clip}=ImageEncoder\left(x_i\right),
\label{eq:s1}
\end{equation}
\begin{equation}
f_j^{text}=TextEncoder\left(t_j\right), j=1,2,3,4,
\label{eq:s2}
\end{equation}
\begin{equation}
S_{ij}=\frac{f_i^{clip}\cdot{{(f}_j^{text})}^T}{||f_i^{clip}||\ {||f}_j^{text}||},
\label{eq:s3}
\end{equation}
\begin{equation}
g_i=\sum_{j=1}^{4}{S_{ij}\times}g_j^{bin}.
\label{eq:s4}
\end{equation}
Obviously, the above simple CLIP-based gaze estimation method does not require any training process, which only relies on the pre-trained CLIP model. We evaluate the performance of this simple attempt on MPIIFaceGaze dataset, whose average angular error is reported in Table~\ref{tab:s2}. 

For comparison purposes, we train a traditional ResNet-based (including ResNet18 and ResNet50) gaze estimation network on ETH-XGaze dataset and test it on MPIIFaceGaze dataset. To be specific, the traditional network takes a facial image as input, extracts a gaze feature of it via several convolutional layers and gets a final gaze via a fully connected layer. From the results we find that even without training on any exclusive gaze estimation datasets, the CLIP achieves performance comparable to ResNet18 and ResNet50 that trained with cross-domain samples. Therefore, we draw a conclusion that \textbf{\textit{the CLIP could understand human gazes subtly, but its potentials have not been fully exploited}}.
\begin{table}[h]
    \centering
    \caption{The correspondences among prototypes, gaze bins and prompts. }
    \begin{tabular}{l @{} c  c }
    \toprule
        Prototype & \begin{tabular}{c@{}}Gaze Bin\\$\left[ \textit{pitch},\textit{yaw}\right]$\end{tabular} & Prompt \\
        \midrule
        up & [0, $\frac{\pi}{2}$] & A photo of a face gazing up. \\
        down & [0, -$\frac{\pi}{2}$] & A photo of a face gazing down. \\
        left & [$\frac{\pi}{2}$, 0] & A photo of a face gazing left. \\
        right & [-$\frac{\pi}{2}$, 0] & A photo of a face gazing right.\\ 
        \bottomrule
    \end{tabular}
    \label{tab:s1}
\end{table}
In this paper, we develop the potentials of the pre-trained CLIP in boosting gaze estimation performance from a real new perspective, which puts CLIP into a supporting role to facilitate the extraction of gaze-related features of a primary gaze estimation network. 
\begin{table}[h]
    \centering
     \caption{The results of ResNet and CLIP on MPIIFaceGaze dataset. We evaulate ResNet18 and ResNet50 on the cross-domain task $D_E  \rightarrow D_M$ which indicates training on ETH-XGaze and testing on MPIIFaceGaze dataset.}
    \begin{tabular}{l c}
    \toprule
    Method & Angular error on MPIIFaceGaze \\
    \midrule
    ResNet18 ($\mathcal{D}_E \rightarrow \mathcal{D}_M$) & $9.40 ^\circ$ \\
    ResNet50 ($\mathcal{D}_E \rightarrow \mathcal{D}_M$) & $12.71 ^\circ$ \\
    CLIP & $11.10 ^\circ$ \\
    \bottomrule
  \end{tabular}
    \label{tab:s2}
\end{table}

\begin{figure*}[t]
\centering
    \includegraphics[width=\linewidth]{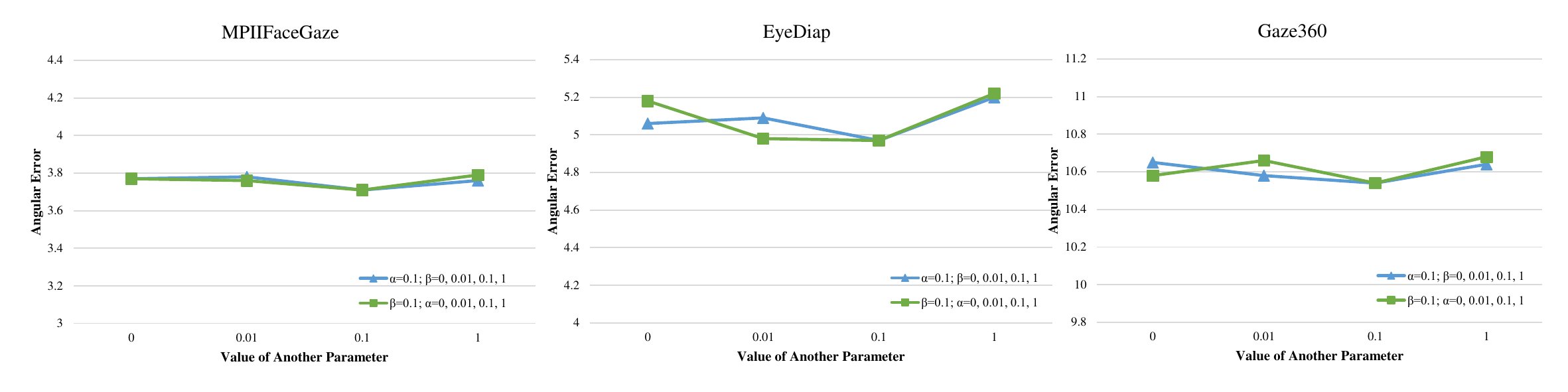}
    \vspace{-10pt}
    \caption{The results of different groups of $\alpha$ and $\beta$ on MPIIFaceGaze, EyeDiap and Gaze360 datasets.}
    \label{fig:s2}
\end{figure*}
\section{Introduction of the datasets used in our experiments}
Our method is evaluated on four popular gaze estimation datasets, which are MPIIFaceGaze, EyeDiap, Gaze360 and ETH-XGaze.

\textit{MPIIFaceGaze}~\cite{MPII} dataset is captured by web cameras in daily life. It contains 45k images from 15 subjects. 

\textit{EyeDiap}~\cite{eye} dataset is collected in laboratory environment with screen targets or floating targets. It contains 16k images of the video clips from 14 subjects.

\textit{Gaze360}~\cite{Gaze360} dataset is captured in both indoor and outdoor environments via a 360° camera. It contains 101k images from 238 subjects with a wide-range head pose and gaze directions.

\textit{ETH-XGaze}~\cite{eth} dataset is collected by high resolution cameras in laboratory environment. It contains 1.1M images from 110 subjects. For our experiments, we select 756,540 images from 80 subjects among them as the training set.

\section{More Discussion}
\subsection{Analyses of hyperparameter}
As discussed in Optimization and Inference, there are three constraints of our DCGaze, which are $ L_{gaze}$, $L_{mask}$ and  $L_{align}$. Their weights are 1, $\alpha$ and $\beta$, respectively. In this section, we discuss the influence of $\alpha$ and $\beta$ in detail. 
We fix $\alpha$ ($\alpha=0.1$) and vary $\beta$ in the range of [0, 0.01, 0.1, 1]. Inversely, we fix $\beta$ ($\beta=0.1$) and vary $\alpha$ in the range [0, 0.01, 0.1, 1]. The experiments are conducted on MPIIFaceGaze, EyeDiap and Gaze360 datasets over the within-domain tasks. As the results shown in Figure~\ref{fig:s2}, there are little variation of angle errors when the hyperparameters $\alpha$, $\beta$ are set to different selected values, which means the DCGaze is insensitive to those hyperparameters.

\subsection{More visualization results}
\begin{figure}[t]
\centering
\begin{subfigure}{\linewidth}
    \includegraphics[width=\linewidth]{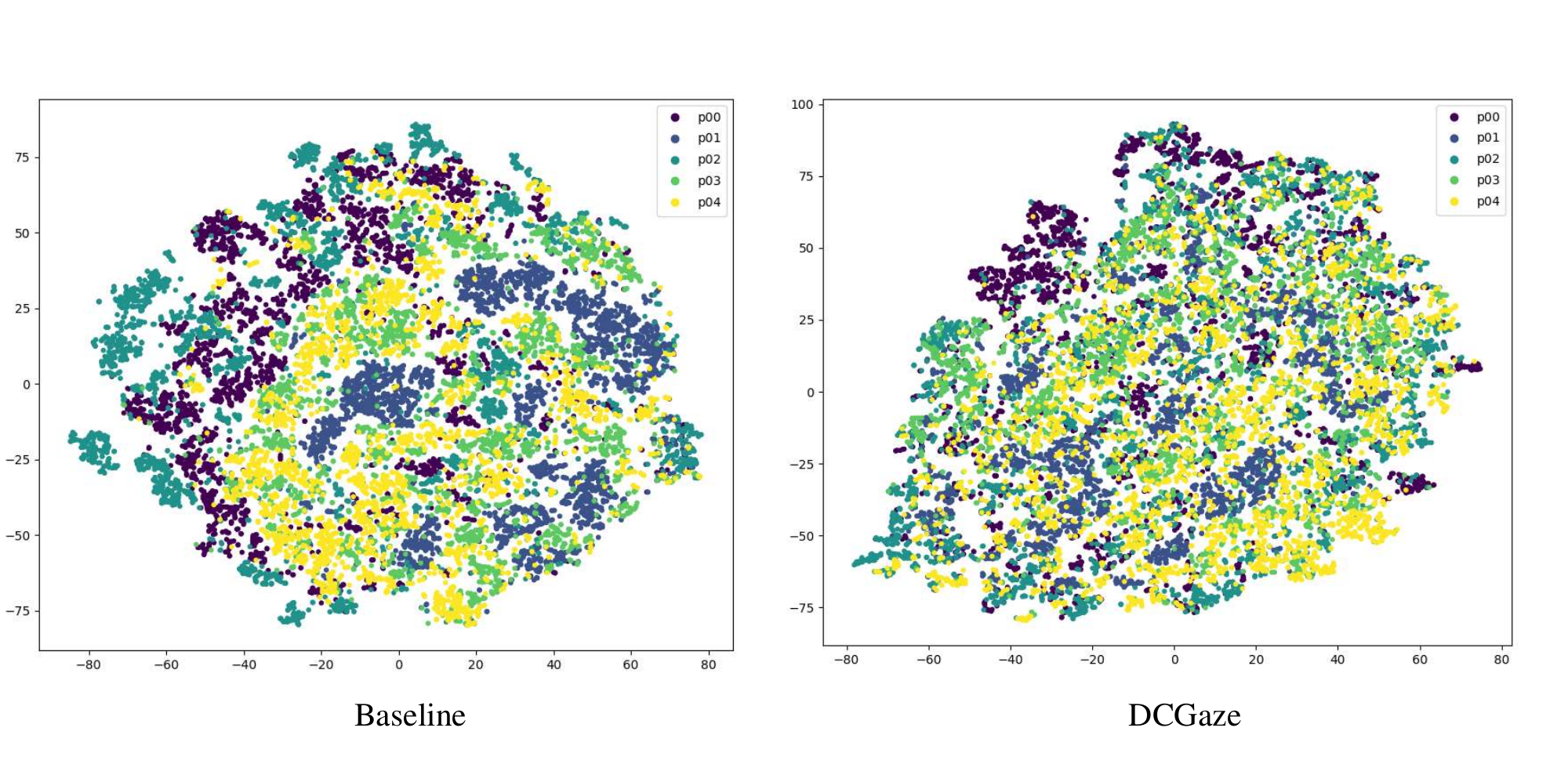}
    \caption{Visualization of features of 5 subjects from MPIIFaceGaze dataset.}
\end{subfigure}
\begin{subfigure}{\linewidth}    \includegraphics[width=\linewidth]{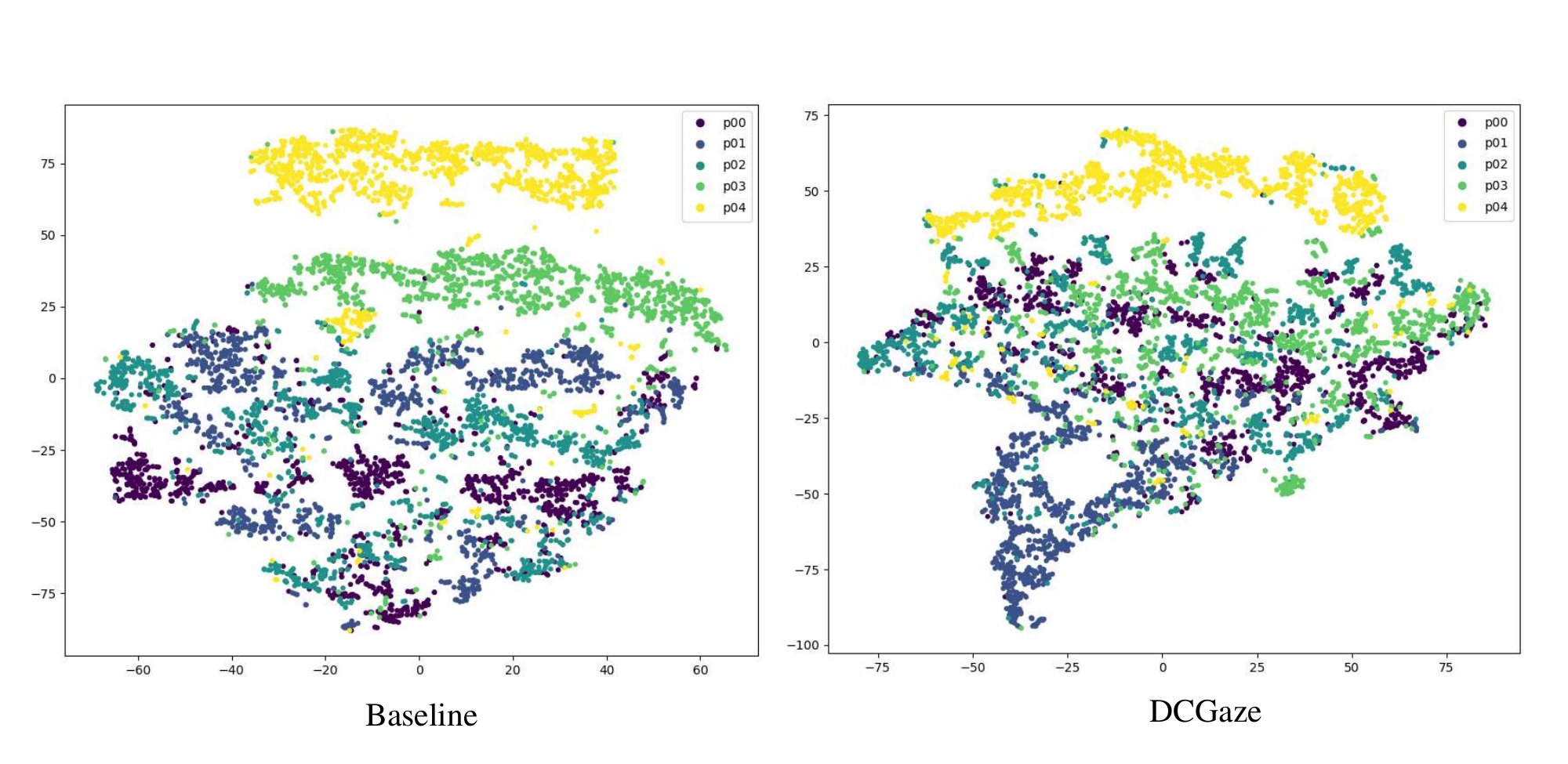}
    \caption{Visualization of features of 5 subjects from EyeDiap dataset.}
\end{subfigure}
 \caption{Visualization of obtained gaze features on MPIIFaceGaze and EyeDiap datasets, where the feature points belonging to the same person share the same color.}
    \label{fig:s4}
\end{figure}
As illustrated in More Discussion, we visualize the distribution of the enhanced features of Gaze360 and ETH-XGaze datasets. In this section, we add more visualization results to demonstrate the advantages of our enhanced gaze features. 

% Firstly, in addition to Gaze360 dataset, we implement the visualization of obtained gaze features on ETH-XGaze dataset. We randomly select 100,000 samples from ETH-XGaze dataset and the results are presented in~\ref{fig:s3}. Similarly, in the scatter plot, the samples are clustered by KMeans~\cite{kmeans} according to their gaze labels, so that the samples with similar gaze directions share similar colors. Obviously, feature points of our enhanced features are distributed in an organized way, in which the features with similar gaze directions are clustered.

We visualize the distributions of gaze features from a subject-specific perspective, in which the feature points belonging to the same person share the same color. We randomly select 5 test subjects from the $D_E\rightarrow D_M$ and $D_E\rightarrow D_D$ tasks and visualize all the gaze features of them obtained via the Baseline and the DCGaze, respectively. Those features are also visualized in 2D using t-SNE~\cite{sne}. The visualization results are shown in Figure~\ref{fig:s4}. It is evident that the features captured by Baseline of each individual are gathered together because they still contain some identity information which is unrelated to gaze. Compared to baseline method, our enhanced features of the same subject are more dispersed, which proves that our DCGaze effectively reduces the variations among subjects and learns purified gaze-related representations.

\subsection{Discussion of scalability of DCGaze}
The number of trainable parameters of our DCGaze and other existing gaze estimation methods are shown in Table~\ref{tab:param1} and Table~\ref{tab:param2} . The scalability of comparison methods is reported in their original papers or counted via reproducing their released codes. In terms of the number of
parameters, our model is comparable to the several lightweight models. Therefore, in the future, we would keep improving DCGaze via reducing its scalability while maintaining the precision of prediction.

\begin{table}[h]
    \caption{Comparisons of scalability with existing methods on within-domain tasks.}
    \centering
    \begin{tabular}{c|ccc|c}
        \toprule
        \multirow{2}{*}{Model} & \multicolumn{3}{c|}{\textbf{Gaze error}} & \textbf{Number of}  \\
        % \cline{2-6}
        \cline{2-4}
         & \textit{MPII} & \textit{EyeDiap} & \textit{Gaze360} & \textbf{ parameters}\\
        \hline
        RT-GENE \cite{Rt}  & 4.30 & 5.90 & - & 82.0M \\
        CA-Net \cite{CA-Net} & 4.37 & 5.27 & 11.20 & 34.1M \\
        PureGaze \cite{PureGaze} & 4.13 & 5.82 & - & 31.55M \\ 
        GazeTR-Pure \cite{GazeTR} & 4.74 & 5.72 & - & 227.3M \\       
        \textbf{DCGaze} & 3.71 & 4.97 & 10.54 & 34.99M \\
        \bottomrule
    \end{tabular}
    \label{tab:param1}
\end{table}

\begin{table}[h]
    \caption{Comparisons of scalability with existing methods on cross domain tasks.}
    \centering
    
    \begin{tabular*}{\linewidth}{c|c@{} c@{} c@{} c|c}
        \toprule
        \multirow{3}{*}{Model} & \multicolumn{4}{c|}{\textbf{Gaze error}} & \multirow{3}{*}{\makecell{\textbf{Number of} \\ \textbf{Parameters}}} \\
        \cline{2-5}
         & \begin{tabular}{c@{}} $\mathcal{D}_E$ \\ $\rightarrow \mathcal{D}_M$ \end{tabular} & \begin{tabular}{c@{}} $\mathcal{D}_E$ \\ $\rightarrow \mathcal{D}_D$ \end{tabular} & \begin{tabular}{c@{}} $\mathcal{D}_G$ \\ $\rightarrow \mathcal{D}_M$ \end{tabular} & \begin{tabular}{c@{}} $\mathcal{D}_G$ \\ $\rightarrow \mathcal{D}_D$ \end{tabular} &  \\
        % \toprule
        % \multirow{2}{*}{Model} & \multicolumn{3}{c|}{\textbf{Gaze error}} & \textbf{Number of}  \\
        % \cline{2-6}
        \cline{2-4}
         % & \textit{MPII} & \textit{EyeDiap} & \textit{Gaze360} & \textbf{ parameters}\\
        \hline
         Gaze360 \cite{Gaze360} & 6.24 & 7.47 & 7.17 & 7.66 & 14.6M \\
        PnP-GA \cite{Pnp-GA} & 5.53 & 5.87 & 6.18 & 7.92 & 116.9M \\
        \textbf{DCGaze} & 6.47 & 6.70 & 6.45  & 7.74 & 34.99M \\
        \bottomrule
    \end{tabular*}
    \label{tab:param2}
\end{table}

\end{document}